\documentclass[final,5p,times,twocolumn,authoryear]{elsarticle}

\usepackage{amssymb}
\usepackage{amsmath}
\usepackage{booktabs}
\usepackage{multirow}
\usepackage{arydshln}
\usepackage[dvipsnames]{xcolor}
\usepackage{subcaption}
\usepackage{graphicx}
\usepackage{algorithm}
\usepackage{algpseudocode}
\usepackage[table]{xcolor}
\usepackage{array}

\usepackage{hyperref}
\hypersetup{
    colorlinks=true,
    linkcolor=blue,
    citecolor=blue,
    urlcolor=blue
}
\journal{Nuclear Physics B}

\begin{document}

\begin{frontmatter}



\title{Controllable Lung Nodule Synthesis via Histogram-Regularized Latent Diffusion Models} 


\author[label1,label2]{Arunkumar Kannan\corref{cor1}}
\ead{akannan7@jhu.edu}

\author[label1]{Yanbo Zhang\corref{cor1}}
\ead{yanbo.zhang@siemens-healthineers.com}

\cortext[cor1]{Corresponding authors}

\author[label1]{Han Liu}
\author[label3]{Michael Baumgartner}
\author[label1]{Jianing Wang}
\author[label1,label4]{Alexander Hertel}
\author[label1]{Bogdan Georgescu}
\author[label1]{Sasa Grbic}

\affiliation[label1]{organization={Digital Technology and Innovation},
            organization={Siemens Healthineers}, 
            city={Princeton},
            state={NJ},
            country={USA}}
\affiliation[label2]{
            organization={Johns Hopkins University}, 
            city={Baltimore}, 
            state={MD},
            country={USA}}
\affiliation[label3]{organization={Digital Technology and Innovation},
            organization={Siemens Healthineers}, 
            city={Erlangen},
            country={Germany}}
\affiliation[label4]{
organization={Department of Radiology and Nuclear Medicine, University Medical Center Mannheim, Heidelberg University},
city={Mannheim},
country={Germany}
}

\begin{abstract}
While automated diagnosis systems have achieved remarkable success in computed tomography (CT)-based lung cancer screening, their development remains limited by the scarcity of diverse,  annotated pulmonary nodule datasets. Diffusion-based generative models offer a promising strategy for data synthesis; however, many existing conditional approaches primarily optimize spatial reconstruction losses, which encourage voxel-wise similarity but may inadequately constrain lesion-level intensity distributions. As a result, these methods may produce over-smoothed texture profiles and underrepresent the distinct attenuation characteristics of different nodule subtypes, including solid, part-solid, and ground-glass nodules. To address this challenge, we propose a controllable latent diffusion model that synthesizes pulmonary nodules within full 3D CT volumes while accurately modeling nodule-specific intensity distributions. Specifically, rather than relying solely on spatial losses, we introduce a histogram-based regularization term that constrains voxel intensity distributions during the generative process. The model combines subtype, spatial mask, and Hounsfield unit (HU) histogram conditioning with the differentiable feature-space histogram regularization term to better align lesion-level intensity distributions, improving the visual plausibility and subtype consistency of synthesized nodules.  Extensive experiments on lung CT data demonstrate that our framework achieves strong visual realism, validated through both quantitative metrics and a visual Turing test. Furthermore, when used for data augmentation, the generated nodules improve performance in downstream clinical tasks, particularly for underrepresented nodule subtypes, and show a potential benefit for subtype-informed malignancy classification.
\end{abstract}






\end{frontmatter}



\section{Introduction}
\label{sec1}

Lung cancer remains the leading cause of cancer-related mortality worldwide, accounting for approximately 1.8 million deaths and nearly 2.5 million new cases annually \citep{bray2024global}. The early detection of pulmonary nodules through low-dose computed tomography (CT) screening is widely recognized as the most effective strategy for improving patient prognosis \citep{national2011reduced}. Major randomized controlled trials, such as the national lung screening trial (NLST) \citep{national2011reduced} and the Dutch-Belgian lung cancer screening trial (Nederlands-Leuvens Longkanker Screenings Onderzoek [NELSON]) \citep{de2020reduced}, have demonstrated that CT screening significantly reduces lung cancer mortality. Consequently, screening programs have been expanded to high-risk populations, with a 2021 study estimating that the eligible screening population in the United States alone exceeds 14 million individuals \citep{landy2021using}. However, the clinical management of detected lung nodules is highly dependent on accurate subtyping. Specifically, pulmonary nodules in CT imaging are categorized based on their attenuation characteristics into three subtypes: (i) solid nodules, characterized by homogeneous soft-tissue attenuation; (ii) ground-glass nodules, non-uniform in appearance with a hazy increase in local attenuation; and (iii) part-solid nodules, comprising both solid and ground-glass attenuation components. This distinction is clinically critical, as subsolid nodules, especially part-solid, frequently exhibit a higher probability of malignancy compared to solid nodules of similar size and often require distinct follow-up protocols or surgical interventions \citep{henschke2002ct, hammer2019cancer}. 

While computer-aided diagnosis, driven by recent advances in deep learning, has emerged as a promising solution to manage expanding screening volumes and support visual subtyping \citep{mortani2026deep, liu2020no,hendrix2023deep}, accurate automated detection and classification of these diverse nodule subtypes remain significant challenges. One primary obstacle is the pronounced ``long-tail'' distribution observed in real-world lung nodule datasets, where the prevalence of biopsy-proven malignant nodules in large screening cohorts is exceedingly low at approximately 3.6\% \citep{national2011national}. A similar imbalance exists at the subtype level, where solid nodules substantially outnumber sub-solid nodules in clinical datasets. Because these high-risk subtypes often exhibit faint, low-contrast textures that are easily overwhelmed by surrounding normal lung tissues and the predominance of benign solid nodules, standard deep learning models trained on such imbalanced data tend to bias toward the majority class. Consequently, these models fail to generalize to the subtle features characteristic of the rare but clinically significant sub-solid nodules. This data scarcity is further exacerbated by high inter-observer variability in radiological interpretation. For example, in the lung image database consortium and image database resource initiative (LIDC-IDRI) study, only 928 out of 2,669 suspected findings achieved consensus among four radiologists \citep{armato2011lung}, making it difficult to curate the large-scale, high-quality annotated datasets required for robust model training. 

To mitigate the scarcity of annotated pathological data, generative modeling has emerged as a promising paradigm for synthetic data augmentation. Early approaches \citep{yang2019class, wang2021realistic} utilized generative adversarial networks (GAN) to synthesize regions of interest (ROI) containing pulmonary nodules, but these often suffered from mode collapse and lacked high-fidelity 3D consistency. More recently, denoising diffusion probabilistic models (DDPM) have demonstrated superior capability in synthesizing realistic pathological features and overcoming these spatial limitations. Foundational frameworks like medical AI for synthetic imaging (MAISI) \citep{guo2025maisi} have employed latent diffusion models (LDM) to generate diverse anatomical structures alongside pathological variations, while models like DiffTumor \citep{chen2024towards} shifted the focus specifically toward generalizable lesion synthesis across abdominal organs. Within the specific domain of pulmonary nodule synthesis, recent frameworks target localized generation. Lesion Diffusion \citep{lei2025lesiondiffusion}, for instance, introduces a text-driven approach to synthesize lesion ROIs and masks from structured reports, whereas LeFusion \citep{lefusion} applies diffusion models directly to voxel space, inpainting synthetic nodules into healthy lung CTs.

However, a trade-off remains between computational scalability and textural fidelity. Methods such as LeFusion achieve state-of-the-art lesion synthesis by incorporating histogram conditioning within a voxel-space diffusion framework, but operating directly in 3D is computationally expensive and restricts these approaches to generating nodules in small, localized regions (e.g., $64 \times 64 \times 32$ crops) rather than full CT volumes. In contrast, LDMs are more efficient, as they operate in a compressed feature space. However, this efficiency comes at the cost of reduced control at the voxel level, often leading to generating over-smoothed textures that fail to capture the subtle characteristics of ground-glass and part-solid lesions.

To address these limitations, we propose a histogram-regularized latent diffusion model (HR-LDM) for 3D pulmonary nodule synthesis. Our approach operates in the compressed latent space, allowing efficient synthesis within full 3D CT volumes rather than isolated local crops. To enforce the explicit voxel-level regularization missing from standard LDMs, we introduce a novel histogram-based regularization term directly into the diffusion objective. This constraint guides the model to learn subtype-specific attenuation distributions, allowing it to better differentiate between solid, part-solid, and ground-glass nodules. As a result, our method combines the scalability of LDMs with the fine-grained textural fidelity of voxel-space approaches. In addition, our framework enables controlled inpainting of diverse nodule subtypes into healthy lung scans, helping mitigate the long-tail distribution present in clinical datasets. To validate our framework, we assess the efficacy of the synthetic data across two clinically relevant downstream tasks: lung nodule subtype and malignancy classification, utilizing both a large-scale internal and a public dataset. Furthermore, we conduct comprehensive quantitative and qualitative evaluations, comparing our generated nodules against state-of-the-art generative frameworks.

The manuscript is organized as follows. Section 2 reviews related work and discusses the limitations of existing nodule synthesis approaches. Section 3 presents our histogram-regularized latent diffusion framework. Section 4 describes the datasets, and Section 5 provides implementation details. Section 6  presents quantitative and qualitative results, including comparisons with state-of-the-art methods. Finally, Section 7 summarizes our contributions and discusses the future outlook of this work.

\section{Related Work}
\label{sec2}

\subsection{Generative Adversarial Networks for Lesion Synthesis}

GANs \citep{goodfellow2020generative} established a foundational framework for synthetic data augmentation in medical imaging, addressing the limited availability of pathological training samples. These approaches have since been adapted to generate a variety of pathological structures, including colon polyps in endoscopy \citep{shin2018abnormal}, liver tumors in CT \citep{frid2018synthetic}, and brain tumors in magnetic resonance imaging \citep{han2018gan}. In the specific domain of pulmonary imaging, early works such as \cite{yang2019class} and \cite{han2019synthesizing} employed conditional GANs to generate ROIs containing nodules, while \cite{jin2021free} explored free-form synthesis to better model irregular lesion shapes.

However, despite their initial success, GAN-based methods frequently suffer from training instability and mode collapse, often yielding outputs with limited diversity that fail to capture the full distribution of real-world pathology. Furthermore, ensuring high-fidelity 3D consistency remains challenging, as standard GAN architectures struggle to reproduce the subtle attenuation profiles required for accurate nodule subtyping. These limitations have motivated the exploration of more stable probabilistic generative models, such as diffusion-based approaches.

\begin{figure*}[!t]
    \centering
    \includegraphics[width=\textwidth]{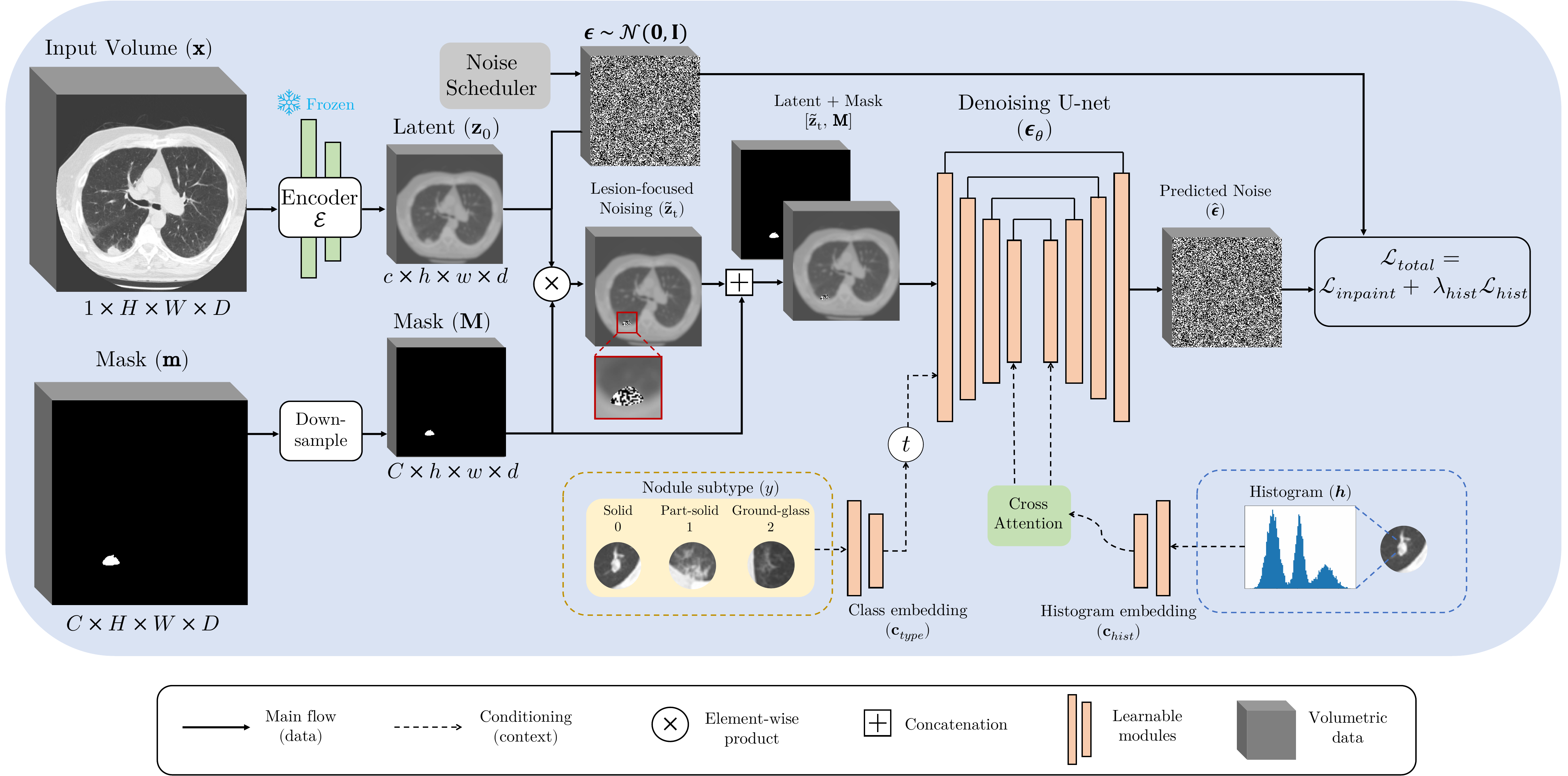}
    \caption{Overview of the proposed framework. The input CT volume is first encoded into a latent representation, where lesion regions are identified using a mask. Noise is added in a lesion-focused manner and the resulting latent is passed, along with the mask, to a denoising U-Net. The model is conditioned on nodule subtype and intensity histogram information through cross-attention. Training is guided by a combination of inpainting and histogram-based losses, allowing realistic reconstruction within lesion regions while preserving overall intensity characteristics.}
    \label{fig:methods}
\end{figure*}

\subsection{Denoising Diffusion Probabilistic Models for Lesion Synthesis}

The emergence of DDPM \citep{ho2020denoising} has shifted the paradigm of medical image synthesis. Unlike GANs, which rely on adversarial training, diffusion models learn to reconstruct data from Gaussian noise via an iterative Markov chain, providing improved training stability and better coverage of the data distribution.

Recent work has explored the use of diffusion models for both large-scale anatomical synthesis and pathology-specific generation. In the context of anatomical synthesis, MAISI \citep{guo2025maisi} represents a significant advancement in scalable 3D medical image generation. MAISI utilizes a high-compression VAE-GAN \citep{esser2021taming,rombach2022high} together with an LDM integrated with ControlNet \citep{zhang2023adding} to generate diverse, high-resolution CT volumes with over 100 controllable anatomical classes.

For pathology synthesis, several specialized frameworks have been proposed. DiffTumor \citep{chen2024towards} introduced a dual-branch architecture to disentangle tumor texture from background context, enabling generalizable synthesis across abdominal organs. More recently, LesionDiffusion \citep{lei2025lesiondiffusion} proposed a text-controllable framework that synthesizes lesion ROIs conditioned on structured radiological reports. These methods typically adopt latent diffusion backbones to manage the computational demands of 3D medical imaging. In contrast, voxel-space approaches such as LeFusion \citep{lefusion} generate lesions directly in the image space and achieve highly detailed outputs by incorporating histogram-based guidance into the diffusion process.

Despite these advances, important trade-offs remain between textural fidelity and computational scalability. Voxel-space methods such as LeFusion provide highly detailed intensity patterns but incur substantial computational cost, limiting synthesis to small localized crops. Latent diffusion approaches, including MAISI and LesionDiffusion, offer substantially improved scalability and enable synthesis over larger volumetric contexts, but they typically rely on coarse semantic conditioning. As a result, they often lack mechanisms to control the intensity patterns that differentiate pulmonary nodule subtypes such as part-solid and ground-glass nodules. These limitations highlight the need for generative frameworks that combine the scalability of latent diffusion with explicit control over lesion intensity characteristics.

\section{Methods}

We propose HR-LDM, a histogram-regularized latent diffusion model for controllable lung nodule generation. Given a healthy CT volume and a set of conditioning signals specifying the desired nodule characteristics, our goal is to synthesize anatomically plausible nodules within the lung parenchyma. 

Figure \ref{fig:methods} provides an overview of the proposed approach, which follows a two-stage generative process. First, a pre-trained volume compression network maps the high-resolution 3D CT volume into a low-dimensional latent space. Next, a DDPM operates in this latent space to inpaint a lung nodule within the CT volume corresponding to the specified subtype. The generation process is guided by three conditioning signals: (i) a \emph{semantic label} that defines the nodule subtype (e.g., solid, ground-glass, or part-solid), (ii) an \emph{intensity histogram} vector that constrains the desired texture profile, and (iii) a \emph{spatial segmentation mask} that defines the nodule location and morphology. 

The remainder of this section describes the architectural components of the framework, including the volume compression network and the lesion-conditioned diffusion training procedure.

\subsection{Stage I: Latent Compression via VAE-GAN}

We employ a pre-trained VAE-GAN model adapted from the MAISI  \citep{guo2025maisi}, which was trained on a large dataset comprising 39,206 3D CT volumes and 18,827 3D MRI volumes. This model maps high-dimensional volumetric data into a compact latent representation. Specifically, an asymmetric VAE encoder maps the input CT volume $\mathbf{x} \in \mathbb{R}^{1 \times H \times W \times D}$ into a latent feature map $\mathbf{z}_0 = \mathcal{E}(\mathbf{x}) \in \mathbb{R}^{c \times h \times w \times d}$. This transformation preserves semantic and textural information while significantly reducing the spatial dimensionality, providing a computationally efficient representation for the subsequent diffusion process. 

The VAE-GAN is trained using a composite objective designed to preserve structural fidelity and local realism. The loss function combines a voxel-wise $L_1$ reconstruction loss, a perceptual loss \citep{zhang2018unreasonable} to maintain perceptual consistency, and a patch-based adversarial loss \citep{yu2022vectorquantized} to enhance texture sharpness. Additionally, a Kullback-Leibler (KL) regularization term constrains the latent distribution toward a standard normal prior, ensuring a well-structured latent space suitable for diffusion modeling.

\subsection{Stage II: Training the Latent Diffusion Model}


Our framework employs a DDPM operating in the compressed latent space produced by the VAE. In this latent space, the diffusion process progressively perturbs the data with Gaussian noise and then learns to reverse this process to generate realistic samples. Formally, let $\mathbf{z}_0 \in \mathbb{R}^{c \times h \times w \times d}$ denote the latent representation of a CT volume. The forward diffusion process is defined as a Markov chain that generates a sequence of noisy latents $\mathbf{z}_1, \dots, \mathbf{z}_T$ by progressively adding Gaussian noise according to a fixed variance schedule $\beta_1, \dots, \beta_T$. At timestep $t$, the noisy latent $\mathbf{z}_t$ can be sampled directly as:

\begin{equation}
    \mathbf{z}_t = \sqrt{\bar{\alpha}_t}\mathbf{z}_0 + \sqrt{1-\bar{\alpha}_t}\boldsymbol{\epsilon}
    \label{eq1}
\end{equation}

\noindent where $\alpha_t = 1 - \beta_t$, $\bar{\alpha}_t = \prod_{s=1}^t \alpha_s$, and $\boldsymbol{\epsilon} \sim \mathcal{N}(\mathbf{0}, \mathbf{I})$. The objective is to learn the reverse diffusion process that progressively denoises $\mathbf{z}_t$ to recover $\mathbf{z}_0$. This reverse process is parameterized by a time-conditional neural network $\boldsymbol{\epsilon}_\theta(\mathbf{z}_t, t, \mathbf{c})$, implemented as a 3D U-net \citep{ronneberger2015u, cciccek20163d}, where $\mathbf{c}$ denotes conditioning inputs. The network is trained to predict the added noise $\boldsymbol{\epsilon}$ by minimizing the simplified objective:

\begin{equation}
    \mathcal{L}_{simple} = \mathbb{E}_{\mathbf{z}_0, \boldsymbol{\epsilon}, t, \mathbf{c}} \left[ \| \boldsymbol{\epsilon} - \boldsymbol{\epsilon}_\theta(\mathbf{z}_t, t, \mathbf{c}) \|_1 \right]
\end{equation}

\noindent Upon convergence, new samples can be generated by iteratively denoising a random Gaussian latent $\mathbf{z}_T \sim \mathcal{N}(\mathbf{0}, \mathbf{I})$, conditioned on $\mathbf{c}$.


A core contribution of our framework is the integration of a histogram-based regularization term into the diffusion training objective. This regularization ensures that the voxel intensities of the synthesized lesion match with its target subtype, which is essential for generating clinically realistic nodules.

The subsequent sections detail the conditioning mechanisms used by the diffusion model, the lesion-conditional inpainting formulation, and the proposed histogram-based regularization strategy.

\subsubsection{Multi-Modal Conditioning Inputs}

We employ a multi-modal conditioning strategy in our framework, where each signal provides complementary information about the nodule to be generated, capturing different aspects of its appearance and structure. Specifically, we condition the noise prediction network $\boldsymbol{\epsilon}_\theta$ on three signals: (i) a semantic subtype label, (ii) an intensity histogram, and (iii) a spatial segmentation mask.

\begin{enumerate}
    \item \textbf{Semantic Label Condition ($\mathbf{c}_{type}$).} We define a discrete nodule subtype label $y \in \{0, \dots, K-1\}$ corresponding to $K$ categories such as solid, part-solid, and ground-glass nodules. Each label is mapped to a dedicated embedding layer, rather than a shared embedding, allowing the model to learn subtype-specific feature representations. Formally, for a given label $y$, its conditioning vector is defined as $\mathbf{c}_{type} = \text{Embed}(y) \in \mathbb{R}^{d_{emb}}$. This vector is added to the timestep embedding $\mathbf{t}_{emb}$ within the U-Net, guiding the denoising process toward the target nodule subtype \citep{nichol2021improved}.

    \item \textbf{Histogram Texture Condition ($\mathbf{c}_{hist}$).} We condition the model on Hounsfield unit (HU) intensity histograms \citep{lefusion} to guide subtype-specific texture generation. Specifically, we represent the intensity distribution of a nodule using a normalized histogram $\mathbf{h} \in \mathbb{R}^{B}$, where $B$ denotes the number of bins. Unlike scalar inputs, such as the subtype label, this vector describes the distribution of voxel intensities within the lesion. We project this histogram using a learnable MLP with SiLU activation:

    \begin{equation}
    \mathbf{c}_{hist} = \text{MLP}(\mathbf{h}) = \mathbf{W}_2 \cdot \sigma(\mathbf{W}_1 \cdot \mathbf{h} + \mathbf{b}_1) + \mathbf{b}_2
    \end{equation}
    
    where $\mathbf{c}_{hist} \in \mathbb{R}^{d_{ctx}}$. This embedding ($d_{ctx}$) is incorporated with the U-Net via cross-attention layers, where it modulates the feature representations based on the target intensity distribution \citep{rombach2022high}.

    \item \textbf{Spatial Mask Condition ($\mathbf{M}$).} We further condition the network using a multi-channel binary segmentation mask of the nodule to constrain its spatial extent and morphology, denoted as $\mathbf{m} \in \{0, 1\}^{C \times H \times W \times D}$, where $C$ represents distinct morphological components (e.g., solid core, ground-glass region, and spiculations). The mask is downsampled via interpolation to match the latent spatial dimensions $(h, w, d)$, resulting in $\mathbf{M} \in \mathbb{R}^{C \times h \times w \times d}$. This downsampled mask is concatenated channel-wise with the noisy latent input $\mathbf{z}_t$, providing an explicit spatial constraint that restricts generation to the nodule region \citep{saharia2022palette,ronneberger2015u}.
\end{enumerate}

\begin{figure*}[!t]
    \centering
    \includegraphics[width=\textwidth]{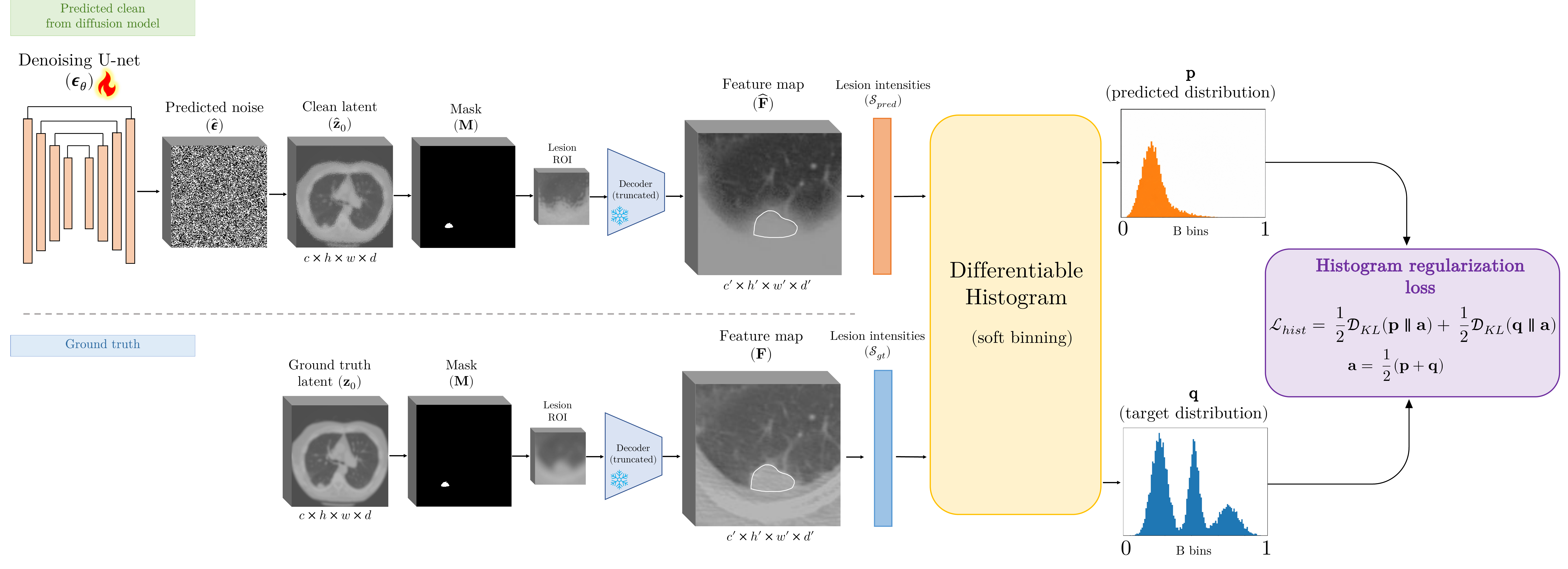}
    \caption{Overview of the proposed histogram-based regularization framework. At each diffusion timestep, the denoising U-Net predicts the noise estimate $\hat{\boldsymbol{\epsilon}}$, which is used to recover the clean latent estimate $\hat{\mathbf{z}}_0$. Instead of decoding to full voxel space, a truncated frozen decoder projects the latent representation into a high-resolution feature space for efficient histogram computation. Feature values inside the lesion region are extracted using the lesion mask and converted into differentiable soft-binned histograms. The predicted histogram distribution $\mathbf{p}$ is then aligned with the corresponding ground-truth distribution $\mathbf{q}$ using a Jensen-Shannon divergence regularization loss.}
    \label{fig:methods-hist}
\end{figure*}

\subsubsection{Lesion-Conditional Inpainting Strategy} 

We formulate the training process as a masked inpainting task \citep{chen2024towards} in the latent space to synthesize lung nodules while preserving the surrounding anatomical context. Unlike standard diffusion approaches where noise is applied globally to the input latent, we construct a composite latent $\tilde{\mathbf{z}}_t$ that combines a clean background with a noisy lesion target. Given a latent volume $\mathbf{z}_0$ and a downsampled binary lesion mask $\mathbf{M}$ (where 0 denotes background and 1 denotes the lesion region), we form $\tilde{\mathbf{z}}_t$ by selectively applying the forward diffusion process within the masked region:

\begin{equation}
    \tilde{\mathbf{z}}_t = \underbrace{(1 - \mathbf{M}) \odot \mathbf{z}_0}_{\text{Clean Background}} + \underbrace{\mathbf{M} \odot \mathbf{z}_t}_{\text{Noisy Lesion}}
\end{equation}

\noindent where $\mathbf{z}_{t}$ is obtained from the forward diffusion process defined in Eq. (\ref{eq1}), and $\odot$ denotes the element-wise Hadamard product. In this formulation, the first term preserves the clean latent representation outside the lesion region, while the second term applies noise according to the diffusion process only within the masked lesion area. Consequently, the model focuses on synthesizing the lesion texture and boundary within the masked region. 

To enforce this constraint during optimization, we modify the standard diffusion objective to calculate the reconstruction error exclusively within the lesion region. Specifically, the noise prediction network $\boldsymbol{\epsilon}_\theta$, which takes as input the concatenated latent and mask $[\tilde{\mathbf{z}}_t, \mathbf{M}]$ along with semantic label $\mathbf{c}_{type}$ and histogram conditioning $\mathbf{c}_{hist}$, is trained using a spatially weighted $L_1$ loss between the predicted noise $\hat{\boldsymbol{\epsilon}}$ (output of $\boldsymbol{\epsilon}_\theta$) and the ground-truth noise $\boldsymbol{\epsilon}$:

\begin{equation}
    \mathcal{L}_{inpaint} = \frac{\sum \| \mathbf{M} \odot (\boldsymbol{\epsilon} - \hat{\boldsymbol{\epsilon}}) \|_1}{\sum \mathbf{M} + \delta}
    \label{spatial}
\end{equation}

\noindent where the summation is over all latent voxels restricted to the lesion region defined by $\mathbf{M}$, and $\delta$ is a small smoothing constant to prevent division by zero. This region-specific loss prevents the model from trivially minimizing the objective by predicting zeros in the background, ensuring that gradients are driven primarily by the nodule synthesis task.

\begin{figure*}[!t]
    \centering
        \includegraphics[width=0.9\textwidth]{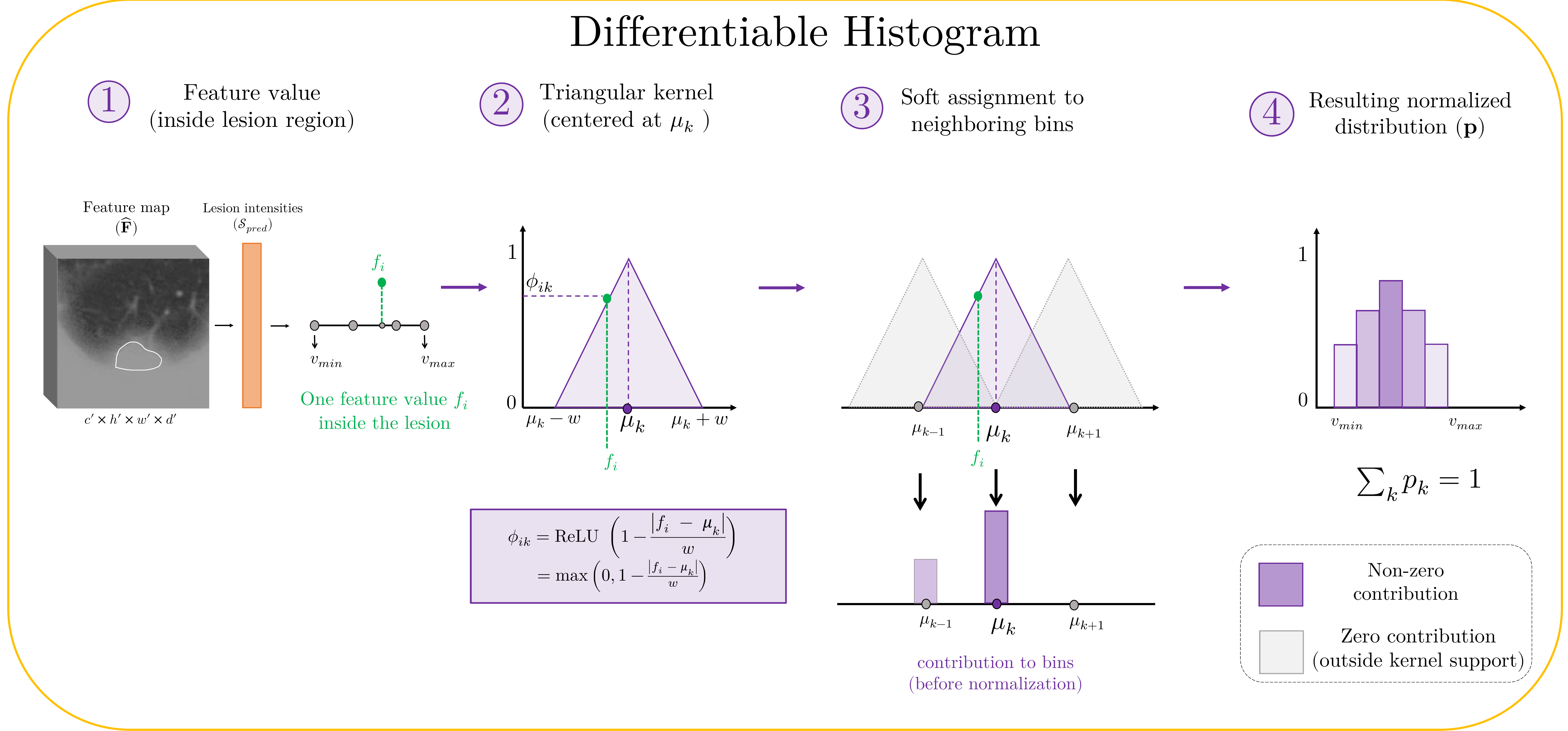}
    
    \caption{Illustration of the differentiable histogram construction used for histogram regularization. Feature values extracted from the lesion region are softly assigned to neighboring histogram bins using a triangular kernel centered at each bin location $\mu_k$. Unlike standard hard-binned histograms, the proposed soft-binning formulation remains differentiable, enabling gradients to propagate back through the histogram computation during diffusion training. The resulting normalized histogram forms the predicted intensity distribution $\mathbf{p}$ used for optimization.}
    \label{fig:methods-hist-loss}
\end{figure*}

\subsubsection{Histogram-Based Regularization}

We introduce a histogram-based regularization term to optimize the intensity distributions of generated pulmonary nodules. Standard reconstruction losses in diffusion models, such as $L_1$ and $L_2$, capture global structure but tend to produce averaged intensity values, leading to over-smoothed or blurry outputs \citep{kingma2013auto}. This limits their ability to represent the distinct attenuation patterns of sub-solid nodules. To address this, we incorporate an intensity-aware constraint into the diffusion training objective. Here, the goal is to align the histogram of the reconstructed lesion with that of the ground-truth lesion during training. However, implementing histogram regularization within LDMs presents two key challenges.

(i) \textbf{Latent space lacks meaningful intensity information.} While histograms are naturally defined in voxel space, latent representations do not retain sufficient intensity information for reliable histogram computation. A direct solution would require projecting the latent representations back into image space at every diffusion timestep, which results in significant computational overhead.

(ii) \textbf{Histogram computation must be differentiable.} Since we propose to use this histogram objective as a part of the optimization process, all operations involved in histogram construction must be differentiable in order to support gradient propagation. However, computing a naive hard-binned histogram is a non-differentiable operation, which prevents gradients from flowing back to the denoising network during training.

Therefore, to address these challenges, we design a histogram regularization framework with two key components: (i) a feature-space approximation strategy that enables efficient histogram computation without repeated full-volume decoding, and (ii) a differentiable soft-binning formulation that preserves end-to-end gradient flow during optimization. The overall histogram regularization pipeline is illustrated in Figure ~\ref{fig:methods-hist}. \\

\textbf{Feature-Space Approximation.} To avoid the computational overhead of decoding latent representations ($\mathbb{R}^{c \times h \times w \times d}$) into full voxel space ($\mathbb{R}^{1 \times H \times W \times D}$) at every diffusion timestep, we compute histogram statistics in an intermediate high-resolution feature representation that retains intensity-related characteristics. Specifically, following \citep{berrada2025boosting}, we use a \emph{truncated} decoder $\mathcal{D}_{trunc}$ that maps the latent to a high-resolution feature space, which serves as a proxy for voxel intensities.

Formally, at timestep $t$ of the diffusion training process, the denoising network predicts the noise $\hat{\boldsymbol{\epsilon}}$. We then recover an estimate of the clean latent volume $\hat{\mathbf{z}}_0$ from the predicted noise using the reverse diffusion formulation \citep{ho2020denoising}:

\begin{equation}
    \hat{\mathbf{z}}_0 = \frac{\mathbf{z}_t - \sqrt{1 - \bar{\alpha}_t} \ \hat{\boldsymbol{\epsilon}}}{\sqrt{\bar{\alpha}_t}}
\end{equation}

\noindent This estimated latent is subsequently passed through a frozen copy of the VAE decoder. Here, instead of using the full decoder pipeline, we truncate the forward pass at the penultimate residual block. At this stage, the feature maps have sufficient spatial resolution to reflect intensity variations while avoiding the significant computational overhead of the final convolution and upsampling layers. We then compute the histogram on the resulting feature map $\hat{\mathbf{F}} = \mathcal{D}_{trunc}(\hat{\mathbf{z}}_0)$ for supervision. \\

\textbf{Differentiable Histogram.} We implement a differentiable soft-binning operation \citep{ustinova2016learning} on the extracted feature map $\hat{\mathbf{F}}$, making it suitable for gradient-based optimization. Instead of standard histogram computation, which relies on non-differentiable hard assignments, we use a continuous approximation to histogram binning. Specifically, we adopt a kernel-based formulation with a triangular kernel to softly assign intensity values to histogram bins. An overview of the differentiable soft-binning process is shown in Figure ~\ref{fig:methods-hist-loss}.

The process is as follows. We first isolate feature values corresponding to the lesion region using the mask $\mathbf{M}$,  resulting in the set $\mathcal{S} = \{ f_i \mid \mathbf{M}_i = 1 \}$. We then construct $B$ histogram bins spanning the range of feature values, with uniformly spaced bin centers $\mu_k$ and a fixed bin width $w$. The contribution of each feature value $f_i$ to the $k$-th bin is computed using a triangular kernel function:

\begin{equation}
    h_k = \frac{1}{|\mathcal{S}|} \sum_{f_i \in \mathcal{S}} \text{ReLU}\left( 1 - \frac{|f_i - \mu_k|}{w} \right)
\end{equation}

\noindent This produces a histogram vector $\mathbf{h} = [h_1, \dots, h_B]$, which we normalize such that $\sum h_k = 1$, to obtain the predicted distribution $\mathbf{p}$. A corresponding target distribution $\mathbf{q}$ is computed in the same manner using features extracted from the ground-truth volume within the lesion region. \\

\textbf{Optimization Objective.} We measure the discrepancy between the predicted distribution $\mathbf{p}$ and the target distribution $\mathbf{q}$ using the Jensen-Shannon divergence (JSD) \citep{lin2002divergence}. Compared to the KL divergence, JSD is symmetric and bounded, and remains numerically stable even when the two-distributions have non-overlapping support. The histogram regularization term is defined as:

\begin{equation}
    \mathcal{L}_{hist} = \frac{1}{2} \mathcal{D}_{KL}\left(\mathbf{p} \parallel \mathbf{a}\right) + \frac{1}{2} \mathcal{D}_{KL}\left(\mathbf{q} \parallel \mathbf{a}\right)
\end{equation}

\noindent where $\mathbf{a} = \frac{1}{2}(\mathbf{p} + \mathbf{q})$ is the average distribution. The final optimization objective for our framework is a weighted combination of the spatial reconstruction loss defined previously in Eq. (\ref{spatial}) and the proposed histogram regularization term:

\begin{equation}
    \mathcal{L}_{total} = \mathcal{L}_{inpaint} + \lambda_{hist} \mathcal{L}_{hist}
\end{equation}

\noindent where $\lambda_{hist}$ is a hyperparameter that balances the trade-off between spatial fidelity and intensity distribution alignment. By minimizing this objective, the model is encouraged to reproduce the specific attenuation profiles of pulmonary nodules (e.g., solid, part-solid, and ground-glass nodules), mitigating the over-smoothing effects commonly observed with purely spatial losses. The overall training procedure is summarized in Algorithm \ref{alg:training}.

\begin{algorithm}[t]
\small
\caption{Training Histogram-Regularized Latent Diffusion Model}
\label{alg:training}
\textbf{Input:} CT volume $\mathbf{x}$, nodule subtype label $y$, intensity histogram $\mathbf{h}$, \\ and lesion mask $\mathbf{M}$

\textbf{while not converged do}
\begin{algorithmic}[1]

\State Encode input: $\mathbf{z}_0 \leftarrow \mathcal{E}(\mathbf{x})$

\State Sample $t \sim \text{Uniform}(1,T)$, $\boldsymbol{\epsilon} \sim \mathcal{N}(\mathbf{0}, \mathbf{I})$

\State Forward diffusion process:
$\mathbf{z}_t = \sqrt{\bar{\alpha}_t}\mathbf{z}_0 + \sqrt{1-\bar{\alpha}_t}\boldsymbol{\epsilon}$
\State Noisy latent (lesion only): $\tilde{\mathbf{z}}_t = (1-\mathbf{M})\odot \mathbf{z}_0 + \mathbf{M}\odot \mathbf{z}_t$

\State Conditioning signals:
$\mathbf{c}_{type} \leftarrow \text{Embed}(y), \ \mathbf{c}_{hist} \leftarrow \text{MLP}(\mathbf{h})$

\State Noise prediction:
$\hat{\boldsymbol{\epsilon}} \leftarrow \boldsymbol{\epsilon}_\theta([\tilde{\mathbf{z}}_t, \mathbf{M}], t, \mathbf{c}_{type}, \mathbf{c}_{hist})$

\State Clean latent: $\hat{\mathbf{z}}_0 = \frac{\mathbf{z}_t - \sqrt{1-\bar{\alpha}_t}\ \hat{\boldsymbol{\epsilon}}}{\sqrt{\bar{\alpha}_t}}$

\State $\mathbf{p} = \text{Hist}(\mathcal{D}_{trunc}(\hat{\mathbf{z}}_0), \mathbf{M}), \;
\mathbf{q} = \text{Hist}(\mathcal{D}_{trunc}(\mathbf{z}_0), \mathbf{M}), \; \mathbf{a} = \frac{1}{2}(\mathbf{p} + \mathbf{q})$

\State \textbf{Inpainting loss:}
$\mathcal{L}_{inpaint} = \frac{\sum \|\mathbf{M}\odot(\boldsymbol{\epsilon}-\hat{\boldsymbol{\epsilon}})\|_1}{\sum \mathbf{M} + \delta}$

\State \textbf{Histogram loss:} $\mathcal{L}_{hist} = \tfrac{1}{2}\mathcal{D}_{KL}(\mathbf{p}\|\mathbf{a}) + \tfrac{1}{2}\mathcal{D}_{KL}(\mathbf{q}\|\mathbf{a})$

\State $\mathcal{L}_{total} = \mathcal{L}_{inpaint} + \lambda_{hist}\mathcal{L}_{hist}$

\State $\theta \leftarrow \theta - \eta \nabla_\theta \mathcal{L}_{total}$

\end{algorithmic}
\end{algorithm}

\subsection{Inference and Controllable Synthesis}

Once trained, our HR-LDM framework can be used to generate synthetic lung nodules in healthy CT scans. At inference time, the model takes a healthy CT scan along with the conditioning signals defined earlier to guide the generation process. Here, we detail the inference process by first describing how anatomically valid locations are identified within the target volume for nodule placement, followed by the construction of conditioning signals (e.g., segmentation masks and attributes), and finally how the generated lesion is seamlessly integrated into the background anatomy.

\begin{enumerate}
    \item \textbf{Candidate Location Selection.} We first identify anatomically valid locations for nodule placement within the target volume using a pre-computed lung lobe segmentation mask obtained with TotalSegmentator \citep{wasserthal2023totalsegmentator,isensee2021nnu}, which localizes the pulmonary parenchyma where nodules can realistically occur. To avoid boundary artifacts (e.g., when part of the nodule lies on the chest wall and appears visually artificial), we define an inner region by applying iterative binary erosion to the lung mask. Centroid locations $\mathbf{c}_{pos}$ for standard parenchymal nodules are then sampled exclusively from this region. To additionally model sub-pleural nodules, we consider the outer region near the lung boundary, defined as the difference between the full lung mask and the eroded inner region. This approach ensures that the synthesized nodules remain consistent with the underlying lung anatomy.

    \item \textbf{Mask and Attribute Conditioning.} Following location selection, we construct conditioning signals that define the nodule morphology and texture. We begin by selecting a target lesion mask $\mathbf{m}_{target}$ from a curated library of real nodule shapes and aligning it spatially with the sampled centroid $\mathbf{c}_{pos}$ within the target volume. We then prepare the conditioning inputs required by the diffusion model, including a semantic label $\mathbf{c}_{type}$ and a target intensity histogram $\mathbf{c}_{hist}$.

    \item \textbf{Latent Inpainting and Reconstruction.} Finally, we perform generation by initializing the reverse diffusion process from Gaussian noise, $\mathbf{z}_T \sim \mathcal{N}(\mathbf{0}, \mathbf{I})$, and iteratively denoising to obtain the final latent $\hat{\mathbf{z}}_0$. Here, at each timestep $t$, the denoising network predicts a latent estimate $\mathbf{z}_{t-1}$ from the current noisy input $\mathbf{z}_t$. Next, consistent with the training process, we condition the denoising network with a spatial mask $\mathbf{M}$, obtained by downsampling the target lesion mask $\mathbf{m}_{target}$. This conditioning guides the model to synthesize content specifically within the lesion region while leaving the surrounding anatomy unchanged. In parallel, to maintain background consistency throughout sampling, we introduce a fixed background latent representation. This is obtained by encoding the input CT volume $\mathbf{x}$ into latent space as $\mathbf{z}^{bg} = \mathcal{E}(\mathbf{x})$, followed by forward diffusion to match the noise level at timestep $t-1$, yielding $\mathbf{z}_{t-1}^{bg}$. The updated latent is then formed by combining the synthesized lesion content with the preserved background:
    
    \begin{equation}
    \mathbf{z}_{t-1} = \mathbf{M} \odot \mathbf{z}_{t-1} + (1 - \mathbf{M}) \odot \mathbf{z}_{t-1}^{bg}
    \end{equation}
    
    This formulation enforces that only the masked region is updated by the generative model, while the background is consistently anchored to the original anatomy at the appropriate noise level. After completing all reverse diffusion steps, the final latent $\hat{\mathbf{z}}_0$ is decoded using the VAE decoder, $\hat{\mathbf{x}} = \mathcal{D}(\hat{\mathbf{z}}_0)$, producing a synthetic CT volume with realistic lesion appearance and well-preserved anatomical context.

\end{enumerate}

The overall generative process is outlined in Algorithm \ref{alg:inference}.

\begin{algorithm}[h!]
\small
\caption{Inference Process}
\label{alg:inference}
\textbf{Input:} Healthy CT volume $\mathbf{x}$, nodule subtype label $y$, intensity histogram $\mathbf{h}$, and lesion mask $\mathbf{m}_{target}$

\begin{algorithmic}[1]
\State Sample centroid $\mathbf{c}_{pos}$ from valid lung region 

\State Spatially align $\mathbf{m}_{target}$ with $\mathbf{c}_{pos}$

\State Conditioning signals:
$\mathbf{c}_{type} \leftarrow \text{Embed}(y), \ \mathbf{c}_{hist} \leftarrow \text{MLP}(\mathbf{h})$ 

\State Encode input: $\mathbf{z}_{bg} \leftarrow \mathcal{E}(\mathbf{x})$

\State Initialize latent: $\mathbf{z}_T \sim \mathcal{N}(\mathbf{0}, \mathbf{I})$

\For{$t = T,\dots,1$}

\State Noise prediction:
$\hat{\boldsymbol{\epsilon}} \leftarrow \boldsymbol{\epsilon}_\theta([\mathbf{z}_t, \mathbf{M}], t, \mathbf{c}_{type}, \mathbf{c}_{hist})$

\State $\mathbf{z}_{t-1} \leftarrow \text{DDPM-step}(\mathbf{z}_t, \hat{\boldsymbol{\epsilon}}, t)$

\State Forward diffusion to background:
$\mathbf{z}_{t-1}^{bg} \leftarrow q(\mathbf{z}^{bg}, t-1)$

\State Latent update:
$\mathbf{z}_{t-1} \leftarrow \mathbf{M}\odot \mathbf{z}_{t-1} + (1-\mathbf{M})\odot \mathbf{z}_{t-1}^{bg}$

\EndFor

\State \textbf{return} $\hat{\mathbf{x}} \leftarrow \mathcal{D}(\hat{\mathbf{z}}_0)$ as generated output.

\end{algorithmic}
\end{algorithm}

\section{Datasets}
In this work, we use a curated collection of 3D lung CT volumes consolidated from multiple sources, including public datasets such as the NLST \citep{nlst2013} and the Duke lung cancer screening dataset (DLCS) \citep{wang2025duke}, along with internal datasets. 

For training the generative model, we utilize nodules from NLST and internal datasets within a clinically relevant size range of 6-30 mm, comprising three subtypes: solid, part-solid, and ground-glass nodules, with 1,838 solid, 365 part-solid, and 747 ground-glass nodules. A held-out set of 390 nodules (243 solid, 32 part-solid, and 115 ground-glass nodules) is used during inference for synthesis, which is subsequently used to augment data for downstream tasks. 

For the lung nodule subtype classification task, the training data follows the same initial distribution of 1,838 solid, 365 part-solid, and 747 ground-glass nodules, where class imbalance is  mitigated through augmentation using synthesized nodules generated by our model. The synthesis process randomizes nodule placement within the CT volume, increasing diversity and improving coverage of underrepresented classes. Evaluation for this subtype classification task is performed on a held-out multi-source dataset aggregated from NLST and internal datasets, consisting of 243 solid, 65 part-solid, and 222 ground-glass nodules. 

Finally, for malignancy classification, we use the DLCS dataset, which includes 2,487 nodules with 2,223 benign and 264 malignant cases. The malignancy labels in DLCS are biopsy-confirmed diagnoses.

\section{Implementation Details}
All models are implemented using PyTorch 2.6.0 and MONAI 1.5.0 \citep{cardoso2022monai}, and trained on NVIDIA V100 and H100 GPUs using CUDA 11.8. The VAE-GAN and LDM components are adopted from the MAISI framework.

For diffusion model training, CT volumes are preprocessed by standardizing channel dimensions, aligning orientations to the RAS coordinate system, and normalizing intensities by clipping to $[-1000, 1000]$ HU and scaling to $[0, 1]$. Spatial dimensions are adjusted to the nearest multiple of 128 to ensure compatibility with the U-Net architecture. The VAE-GAN compresses the input by a factor of 4 to obtain latent representations.

The diffusion model is conditioned on nodule subtype, segmentation masks, and intensity histograms. Subtype labels are encoded as 0 (solid), 1 (part-solid), and 2 (ground-glass). Segmentation masks, annotated at a fine-grained level by radiologists, capture different nodule components (e.g., solid, ground-glass, and spiculations) and are downsampled by a factor of 4 to match the latent resolution before being concatenated with the latent input. Intensity histograms are computed by clipping values to $[-1000, 400]$ HU, discretizing into 40 bins, and normalized to form a probability distribution.

The diffusion model is trained for 200 epochs with a batch size of 1 and an initial learning rate of $1 \times 10^{-5}$. A U-Net is used for noise prediction, with optimization performed using Adam and a polynomial learning rate scheduler. Noise is added through a predefined DDPM scheduler with 1000 training timesteps using a scaled linear beta schedule, where the noise variance parameters increase from $\beta_{start}=0.0015$ to $\beta_{end}=0.0195$, and the model is trained using an L1 loss with histogram regularization (weight = 10.0). Mixed precision training and gradient scaling are used to improve computational efficiency. 

For lung nodule subtype classification, we use a 3D DenseNet-121 model \citep{huang2017densely}. Input volumes are cropped to $90 \times 90 \times 90$, resampled to $1 \times 1 \times 1$ mm resolution, and intensity-normalized in the same manner as the diffusion model. The model is trained for 100 epochs with a batch size of 2 and a learning rate of 
 $1 \times 10^{-4}$, using an 80/20 train–validation split with early stopping. Cross-entropy loss is used for optimization. When synthetic samples are included during training, the validation set contains only real samples.

For malignancy classification, we train three variants of 3D DenseNet-121 with a modified binary output layer. The first is trained from scratch on the DLCS dataset. The second and third are model weights initialized from subtype classification models trained on imbalanced and balanced data, respectively, and fine-tuned for malignancy prediction. This initialization is motivated by the relationship between nodule subtype and malignancy, as malignant nodules often present as part-solid or ground-glass lesions, allowing subtype pretraining to provide useful feature representations.

\section{Results}

We evaluate the proposed framework in terms of both the quality of synthesized nodules and their impact on downstream tasks. We first assess the realism of the generated lesions using quantitative metrics and visual inspection. We then examine their effect on lung nodule subtype classification, followed by an analysis under limited training data, and finally evaluate performance on malignancy classification.

\subsection{Quantitative Evaluation of Synthetic Lesions}

\begin{table*}
\centering
\caption{Quantitative evaluation of lung nodule synthesis across different generative methods. Our proposed histogram-regularized latent diffusion model achieves the lowest distribution distances across all anatomical planes. Lower $(\downarrow)$ is better for all metrics. Best results are highlighted in bold. FID: Fréchet Inception Distance; KID: Kernel Inception Distance; MMD: Maximum Mean Discrepancy.}
\label{tab:quant_results}
\resizebox{\textwidth}{!}{%
\setlength{\tabcolsep}{4pt}
\renewcommand{\arraystretch}{1.4}
\begin{tabular}{lcccccccccccc}
\toprule
& \multicolumn{4}{c}{\textbf{FID} $(\downarrow)$} 
& \multicolumn{4}{c}{\textbf{KID} $(\downarrow) \times 10^{-3}$} 
& \multicolumn{4}{c}{\textbf{MMD} $(\downarrow) \times 10^{-2}$} \\
\cmidrule(lr){2-5} \cmidrule(lr){6-9} \cmidrule(lr){10-13}
\textbf{Method} 
& Axial & Sagittal & Coronal & Avg 
& Axial & Sagittal & Coronal & Avg 
& Axial & Sagittal & Coronal & Avg \\
\midrule\midrule

Diff-Tumor        & 8.4120 & 8.7503 & 9.1055 & 8.7559 & 6.8431 & 7.2147 & 7.9562 & 7.3380 & 2.6418 & 2.8126 & 3.0574 & 2.8373  \\
LesionDiffusion   & 5.6214 & 5.8033 & 6.2104 & 5.8784 & 3.6125 & 4.0381 & 4.5637 & 4.0714 & 1.5832 & 1.6679 & 1.8295 & 1.6935 \\
MAISI (Rectified flow)  & 3.8401 & 3.9115 & 4.1560 & 3.9692 & 1.9348 & 2.1186 & 2.3472 & 2.1335 & 0.9587 & 1.0284 & 1.1469 & 1.0447 \\
MAISI-DDPM (no $\mathcal{L}_{hist}$) & 3.1054 & 3.2201 & 3.4588 & 3.2614 & 1.3129 & 1.4287 & 1.6473 & 1.4630 & 0.7264 & 0.7895 & 0.8462 & 0.7807 \\
\textbf{Ours}     & \textbf{2.8192} & \textbf{2.8400} &  \textbf{3.0308} & \textbf{2.8967} & \textbf{1.0423} & \textbf{1.1189} & \textbf{1.1365} & \textbf{1.0992} & \textbf{0.5568} & \textbf{0.6671} & \textbf{0.7284} & \textbf{0.6507}   \\

\bottomrule
\end{tabular}%
}
\end{table*}

We evaluate the quality of synthesized lung nodules by comparing our method against several recent approaches, including LesionDiffusion \citep{lei2025lesiondiffusion}, Diff-Tumor \citep{chen2024towards}, and variants of MAISI \citep{guo2025maisi}, including the rectified-flow formulation \citep{zhao2026maisi} and a DDPM-based variant without histogram regularization. We quantify distributional similarity between real and synthetic samples using three complementary metrics: Fréchet inception distance (FID), kernel inception distance (KID), and maximum mean discrepancy (MMD). 

We compute these metrics on a subset of 500 real and 500 synthetic nodule volumes, with approximately uniform representation across different nodule subtypes. For this evaluation, we extract feature representations from the nodules using a ResNet-50 model pretrained on RadImageNet \citep{mei2022radimagenet}, which is designed for radiological data. Each 3D CT volume is decomposed into axial, sagittal, and coronal views, and multiple slices around the lesion center are sampled. These slices are passed through the network, and deep feature embeddings are obtained via global average pooling. The resulting feature distributions for real and synthesized nodules are used to compute FID, KID, and MMD. Since lung nodules occupy only a small portion of the full CT volume, metrics computed over the entire image would be dominated by background anatomy and fail to capture differences in nodule quality. Therefore, we perform all evaluations on cropped regions centered around the lesions, so that the metrics reflect the fidelity of the synthesized nodules.

\begin{figure*}[!t]
    \centering
    \includegraphics[width=\textwidth]{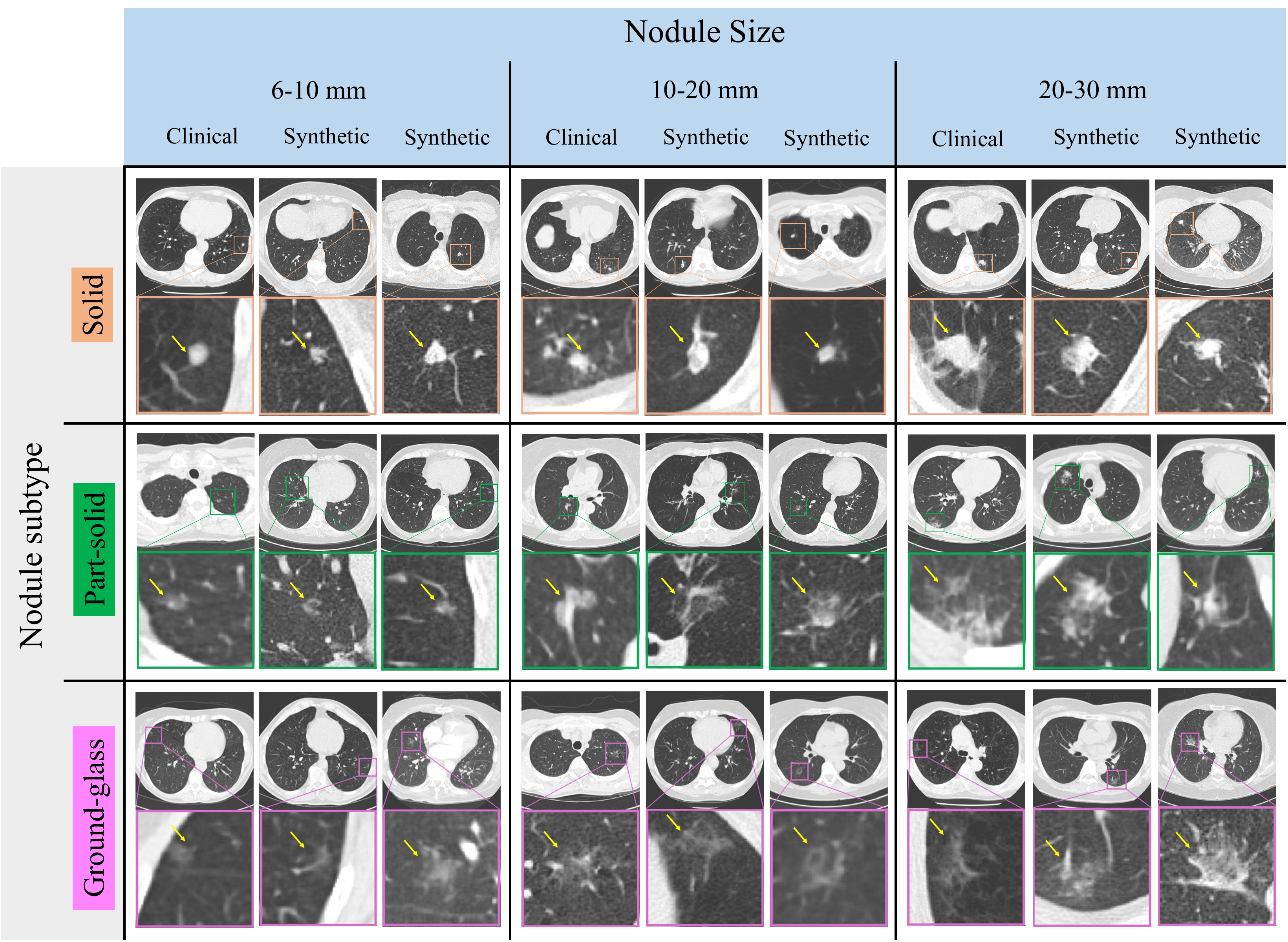}
    \caption{Qualitative comparison of clinical and synthetic pulmonary nodules across varying sizes and nodule types. Each row corresponds to a nodule type (solid, part-solid, and ground-glass), and each column denotes a size range (6-10 mm, 10-20 mm, and 20-30 mm). For each setting, one clinical example and two synthetic samples are shown, along with corresponding zoomed-in views highlighting local morphological details (yellow arrows).}
    \label{qualitative-all}
\end{figure*}

The quantitative results are summarized in Table~\ref{tab:quant_results}. Our method achieves lower FID, KID, and MMD scores across all anatomical planes, as well as on average, compared to the competing methods. This consistent improvement indicates that the distribution of synthesized nodules from our model more closely matches that of real nodules, suggesting better realism and fidelity. We also include qualitative comparisons between clinical and synthetic nodules across different sizes and subtypes in Figure \ref{qualitative-all}.

\subsection{Visual Turing Test}

\renewcommand{\arraystretch}{1.3}

\begin{table}[h!]
\centering
\caption{Visual Turing test results assessing the visual fidelity of synthetic lung nodules. An expert radiologist evaluated 50 real and 50 synthetic nodules to classify them as `Real' or `Synthetic'. The evaluation was performed under two conditions: `Cropped' (assessing isolated nodule morphology) and `Full Context' (assessing anatomical consistency within the surrounding lung volume). Rows denote the ground truth and columns denote the radiologist's predictions.}
\label{tab:turing_confusion}
\setlength{\tabcolsep}{6pt}
\newcolumntype{C}{>{\centering\arraybackslash}m{1.2cm}}
\begin{tabular}{lCCCC}
\toprule
& \multicolumn{4}{c}{\textbf{Radiologist Evaluation}} \\
 \cmidrule(lr){2-5}
& \multicolumn{2}{c}{\textbf{Cropped}} & \multicolumn{2}{c}{\textbf{Full Context}} \\
\cmidrule(lr){2-3} \cmidrule(lr){4-5}
\textbf{Ground Truth} & Real & Synth. & Real & Synth. \\
\midrule
\textbf{Real} 
& 36 & 14 
& 28 & 22 \\

\textbf{Synth.} 
& 20 & 30 
& 22 & 28 \\
\bottomrule
\end{tabular}
\end{table}

To qualitatively assess the realism of the synthetic nodules generated by our framework, we conducted a Visual Turing test with an expert radiologist. The evaluation comprised a blind dataset of 100 cases (50 real clinical and 50 synthetic nodules, comprising a mixture of solid, part-solid and ground-glass). This test was conducted in two phases to assess both the intrinsic appearance of the nodules and their consistency with surrounding anatomy: first, by evaluating cropped nodules, and second, by assessing full-context lung volumes.

In the first phase, the radiologist evaluated tightly cropped nodules, isolated from their surrounding anatomical context, to assess morphological and textural realism; we show in Table \ref{tab:turing_confusion} that this resulted in an overall accuracy of 66.0\% (66/100). Specifically, real nodules were correctly identified with 72.0\% accuracy (36/50), while synthetic nodules were identified with 60.0\% accuracy (30/50). Notably, the synthetic nodules successfully deceived the expert in 40.0\% of the cases (20/50 labeled as real), indicating a high degree of visual fidelity. In addition, the radiologist misclassified 28.0\% (14/50) of the real nodules as synthetic, highlighting the difficulty of distinguishing true clinical data when local contextual cues are removed.

In the second phase, the radiologist evaluated nodules within full-context lung volumes to determine if anatomical placement or boundary inconsistencies revealed their synthetic nature. Interestingly, the addition of contextual information decreased the radiologist's overall accuracy to 56.0\% (56/100), approaching random chance. Under these full-volume conditions, the accuracy for identifying real nodules dropped to 56.0\% (28/50), while the accuracy for synthetic nodules was 56.0\% (28/50). Consequently, the synthetic fooling rate increased to 44.0\% (22/50), while the misclassification of real data as synthetic rose to 44.0\% (22/50).

\begin{figure}
    \centering
    \includegraphics[width=\linewidth]{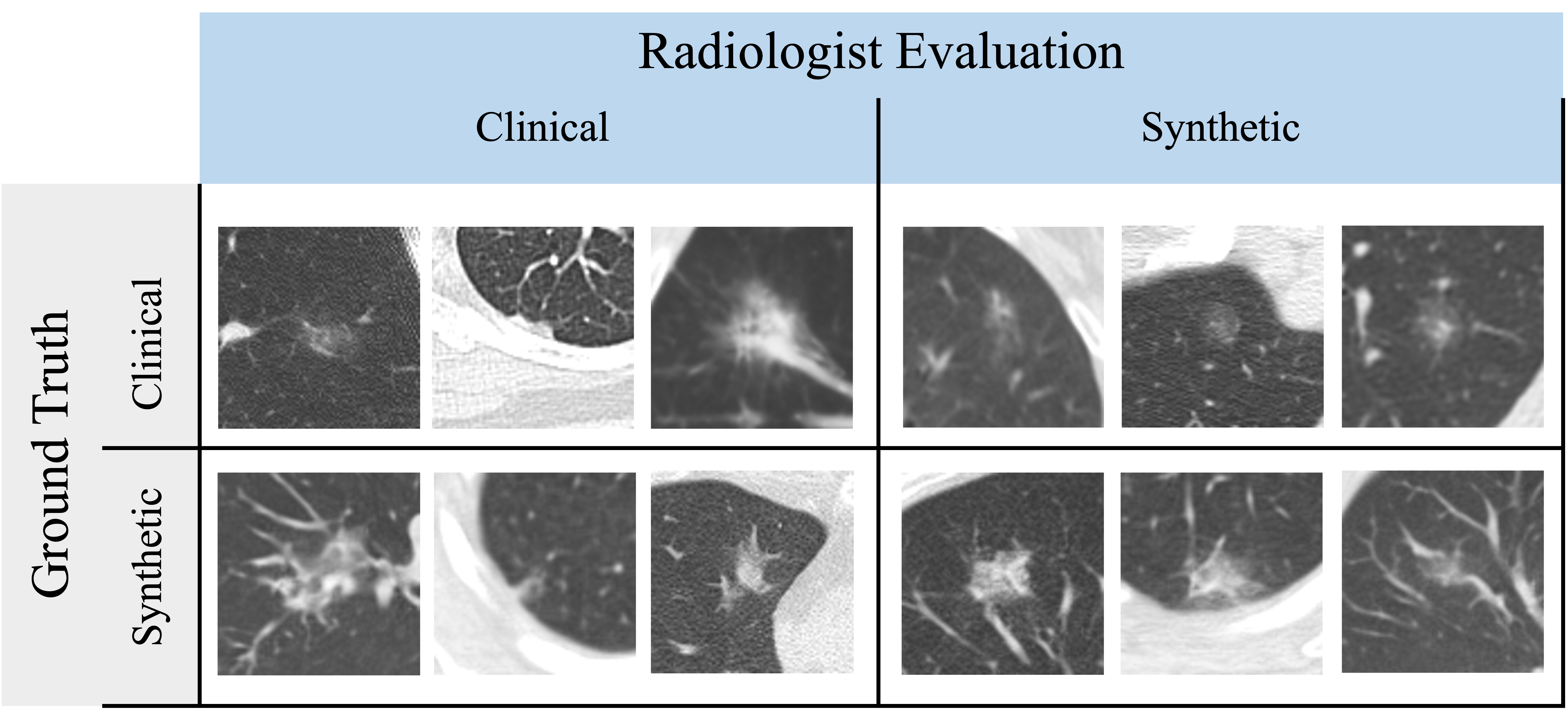}
    \caption{Subjective radiologist evaluation of clinical and synthetic images. Example patches are arranged in a confusion-matrix layout, where rows correspond to the ground-truth label (clinical or synthetic) and columns indicate the radiologist’s assessment. Diagonal entries show correctly identified samples, while off-diagonal entries highlight cases where synthetic images are evaluated as clinical ones (and vice versa). These examples provide a qualitative view of the perceptual similarity between the generated and real images.}
    \label{fig:placeholder}
\end{figure}

These results demonstrate that our proposed method generates synthetic nodules that are often difficult to distinguish from real lesions to an expert observer. The drop in performance from the cropped to the full-context evaluation is particularly notable. This indicates that the model is not only capturing realistic nodule appearance at the local level, but is also placing them in anatomically plausible locations. As a result, the generated nodules remain consistent with the surrounding structures, making them harder to identify even when full context is available. Representative examples from the subjective evaluation are shown in Figure  \ref{fig:placeholder}, where both correctly classified and misclassified cases are presented. The observed confusion between clinical and synthetic nodules further highlights the perceptual similarity between generated and real lesions.

\subsection{Nodule Subtype Classification}

\begin{table}[htbp]
\centering
\scriptsize
\renewcommand{\arraystretch}{0.9}
\caption{Lung nodule subtype classification performance demonstrating the efficacy of diffusion-based data augmentation. The table compares a baseline classifier trained exclusively on the imbalanced real dataset (`Real only') against our method, which leverages synthetic nodules to balance the training distribution (`Real + Synth'), with performance stratified by nodule size and morphological subtype. Metrics evaluated are area under the receiver operating characteristic curve (AUROC) and area under the precision-recall curve (AUPRC), where higher values ($\uparrow$) indicate better performance; subscript values in \textcolor{Gray}{gray} represent standard deviations, \textbf{bold} text denotes superior performance by the augmented model, and markers indicate statistical significance of the improvement over the baseline (\underline{underline}: $p < 0.05$; $*$: $p < 0.01$)}
\label{tab:subtype-classification}
\begin{tabular}{llccc}
\toprule
\textbf{Size} & \textbf{Nodule Type} & \textbf{Baseline} & \textbf{Ours} & \textbf{p-value} \\
& & \textit{Real only} & \textit{Real + Synth} & \\
\midrule
\multicolumn{5}{c}{AUROC ($\uparrow$)} \\

\multirow{3}{*}{\textbf{6-10mm}} 
 & Solid      & 0.977 $\pm$ \scriptsize\textcolor{Gray}{0.01} & 0.971 $\pm$ \scriptsize\textcolor{Gray}{0.01} & 0.910 \\
 & Part-solid & 0.792 $\pm$ \scriptsize\textcolor{Gray}{0.03} & 0.773 $\pm$ \scriptsize\textcolor{Gray}{0.04} & 0.705 \\
 & Ground-glass        & 0.964 $\pm$ \scriptsize\textcolor{Gray}{0.01} & 0.965 $\pm$ \scriptsize\textcolor{Gray}{0.01} & 0.590 \\
\cmidrule{1-5}
\multirow{3}{*}{\textbf{10-20mm}} 
 & Solid      & 0.974 $\pm$ \scriptsize\textcolor{Gray}{0.01} & 0.985 $\pm$ \scriptsize\textcolor{Gray}{0.01} & 0.595 \\
 & Part-solid & 0.753 $\pm$ \scriptsize\textcolor{Gray}{0.04} & \textbf{0.802} $\pm$ \scriptsize\textcolor{Gray}{0.03} & \underline{0.047} \\
 & Ground-glass        & 0.942 $\pm$ \scriptsize\textcolor{Gray}{0.01} & \textbf{0.953} $\pm$ \scriptsize\textcolor{Gray}{0.01} & \underline{0.046} \\
\cmidrule{1-5}
\multirow{3}{*}{\textbf{20-30mm}} 
 & Solid      & 0.979 $\pm$ \scriptsize\textcolor{Gray}{0.02} & 0.986 $\pm$ \scriptsize\textcolor{Gray}{0.01} & 0.373 \\
 & Part-solid & 0.750 $\pm$ \scriptsize\textcolor{Gray}{0.03} & \textbf{0.831} $\pm$ \scriptsize\textcolor{Gray}{0.03} & 0.006$^{*}$ \\
 & Ground-glass        & 0.887 $\pm$ \scriptsize\textcolor{Gray}{0.02} & \textbf{0.933} $\pm$ \scriptsize\textcolor{Gray}{0.02} & 0.004$^{*}$ \\

\midrule
\midrule

\multicolumn{5}{c}{AUPRC ($\uparrow$)} \\

\multirow{3}{*}{\textbf{6-10mm}} 
 & Solid      & 0.961 $\pm$ \scriptsize\textcolor{Gray}{0.01} & 0.955 $\pm$ \scriptsize\textcolor{Gray}{0.03} & 0.953 \\
 & Part-solid & 0.332 $\pm$ \scriptsize\textcolor{Gray}{0.06} & \textbf{0.373} $\pm$ \scriptsize\textcolor{Gray}{0.05} & \underline{0.049} \\
 & Ground-glass        & 0.955 $\pm$ \scriptsize\textcolor{Gray}{0.01} & 0.956 $\pm$ \scriptsize\textcolor{Gray}{0.01} & 0.590 \\
\cmidrule{1-5}
\multirow{3}{*}{\textbf{10-20mm}} 
 & Solid      & 0.894 $\pm$ \scriptsize\textcolor{Gray}{0.03} & 0.901 $\pm$ \scriptsize\textcolor{Gray}{0.04} & 0.590 \\
 & Part-solid & 0.493 $\pm$ \scriptsize\textcolor{Gray}{0.07} & \textbf{0.570} $\pm$ \scriptsize\textcolor{Gray}{0.03} & \underline{0.032} \\
 & Ground-glass        & 0.952 $\pm$ \scriptsize\textcolor{Gray}{0.02} & \textbf{0.962} $\pm$ \scriptsize\textcolor{Gray}{0.01} & \underline{0.046} \\
\cmidrule{1-5}
\multirow{3}{*}{\textbf{20-30mm}} 
 & Solid      & 0.960 $\pm$ \scriptsize\textcolor{Gray}{0.03} & 0.974 $\pm$ \scriptsize\textcolor{Gray}{0.01} & 0.343 \\
 & Part-solid & 0.532 $\pm$ \scriptsize\textcolor{Gray}{0.04} & \textbf{0.657} $\pm$ \scriptsize\textcolor{Gray}{0.05} & 0.002$^{*}$ \\
 & Ground-glass        & 0.848 $\pm$ \scriptsize\textcolor{Gray}{0.03} & \textbf{0.910} $\pm$ \scriptsize\textcolor{Gray}{0.03} & 0.013$^{*}$ \\
\bottomrule
\end{tabular}%
\end{table}

In this section, we evaluate the performance of our approach for lung nodule subtype classification task. The evaluation compares a baseline model, particularly 3D DenseNet-121, trained exclusively on real, imbalanced data, against the model trained using real data augmented with synthetic cases generated by our method. Here, we balance the minority classes (part-solid and ground-glass nodules) in the training dataset with synthetic cases to match the solid nodule counts. We quantify classification performance using both the area under the receiver operating characteristic curve (AUROC) and the area under the precision-recall curve (AUPRC) on a held-out test set consisting only of real cases. To assess statistical significance, we perform a two-sided Mann-Whitney U test comparing metric distributions across multiple independent runs. We use five-fold cross-validation on the training/validation set for model selection, and evaluate the selected models on the test set. Table \ref{tab:subtype-classification} shows the performance of the model stratified by nodule size (6-10mm, 10-20mm, and 20-30mm) and subtype (solid, part-solid, and ground-glass).

The baseline model achieved robust performance for the majority solid nodule class across all size categories. After augmenting with the synthetic data, the model maintained this high performance, with AUROC remaining above 0.97 and AUPRC above 0.90 across all sizes. Furthermore, we also note that the differences in model performance for solid nodules were not statistically significant ($p > 0.05$), indicating that our augmentation strategy did not degrade the representation or classification accuracy of the well-represented majority class.

\begin{figure}[!t]
    \centering
    \begin{subfigure}[b]{0.48\linewidth}
        \centering
        \includegraphics[width=\linewidth]{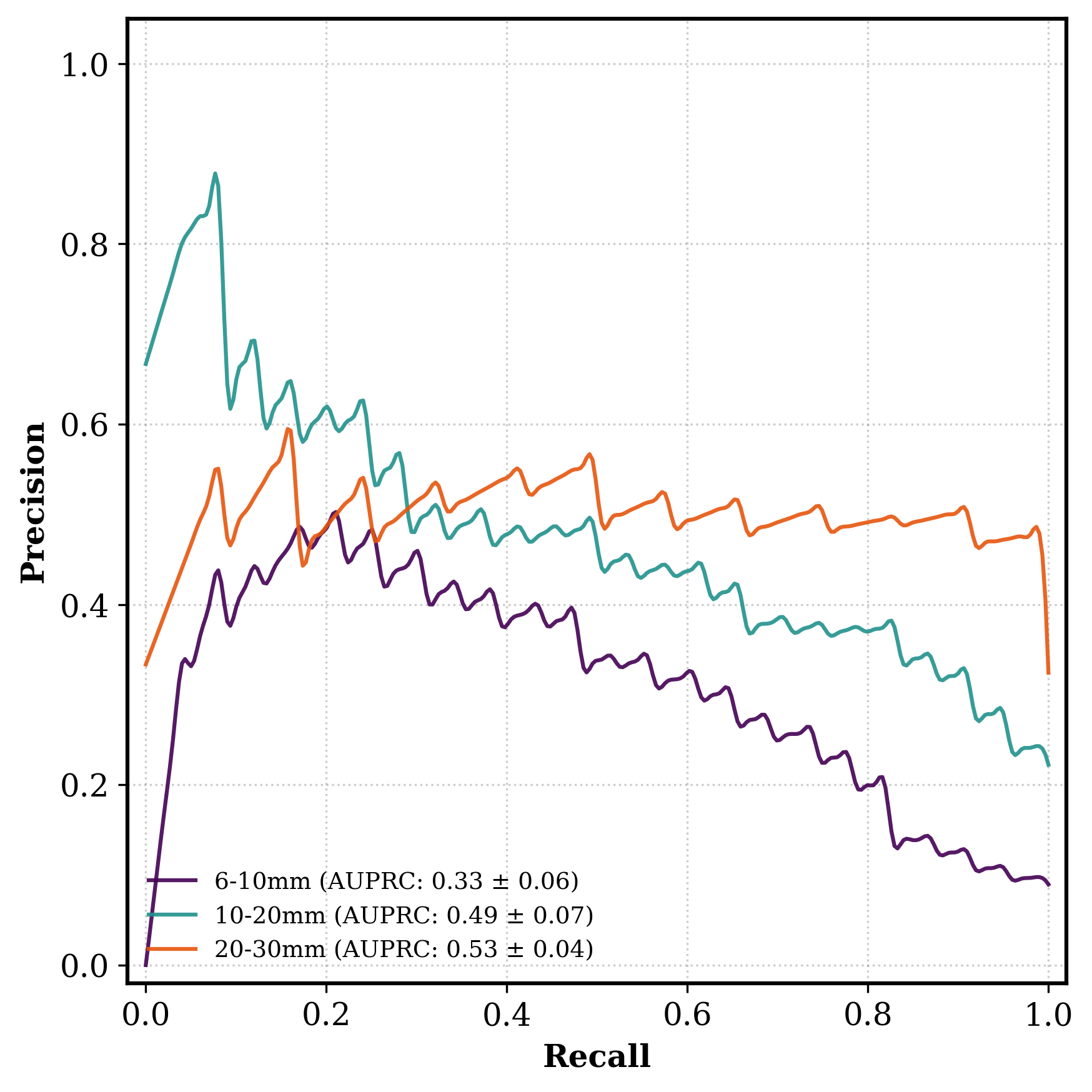} 
        \caption{Part-solid: \textit{Real only} (Baseline)}
        \label{fig:ps-imbalanced}
    \end{subfigure}
    \hfill 
    \begin{subfigure}[b]{0.48\linewidth}
        \centering
        \includegraphics[width=\linewidth]{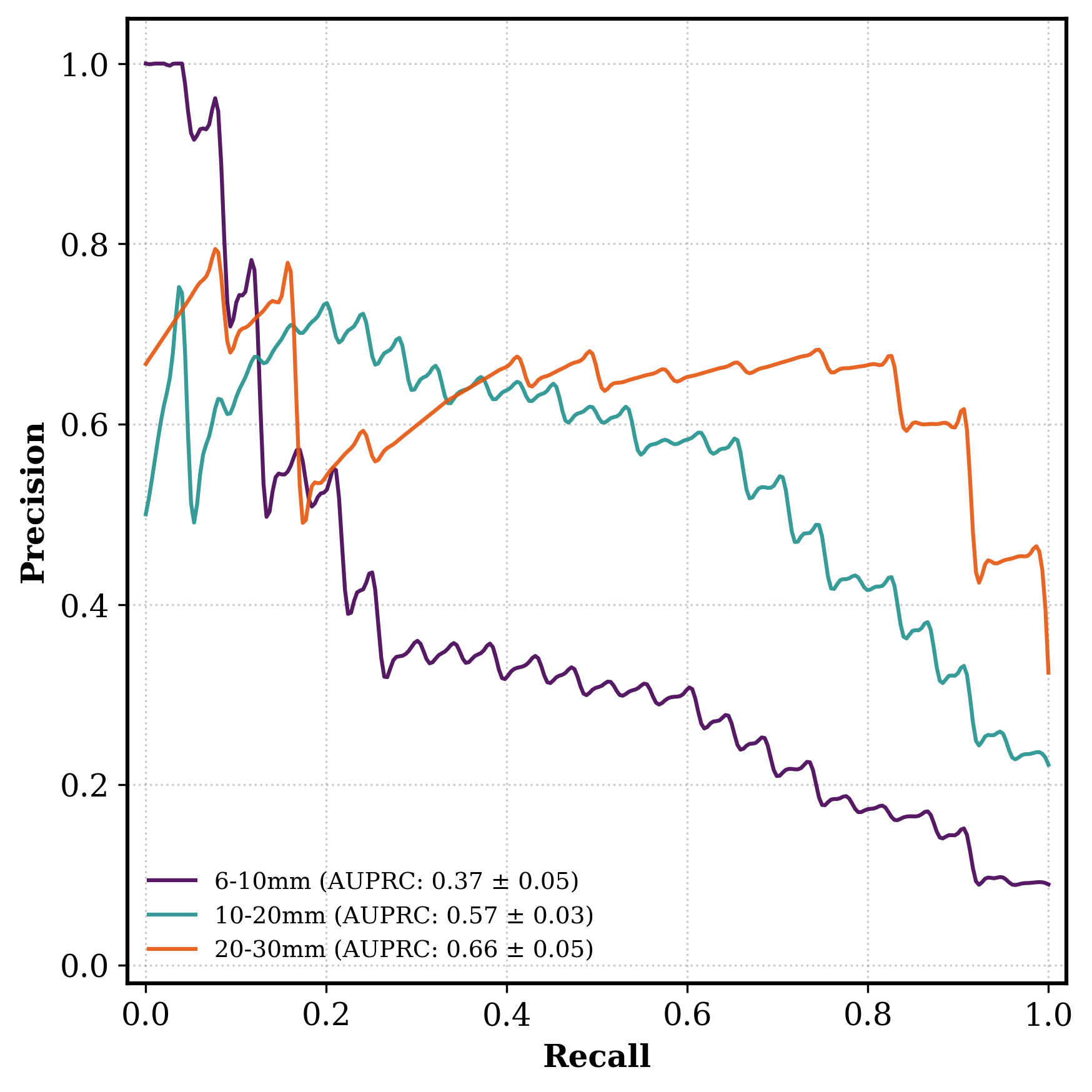} 
        \caption{Part-solid: \textit{Real + Synth} (Ours)}
        \label{fig:ps-diffusion}
    \end{subfigure}
    
    \vspace{1em} 

    \begin{subfigure}[b]{0.48\linewidth}
        \centering
        \includegraphics[width=\linewidth]{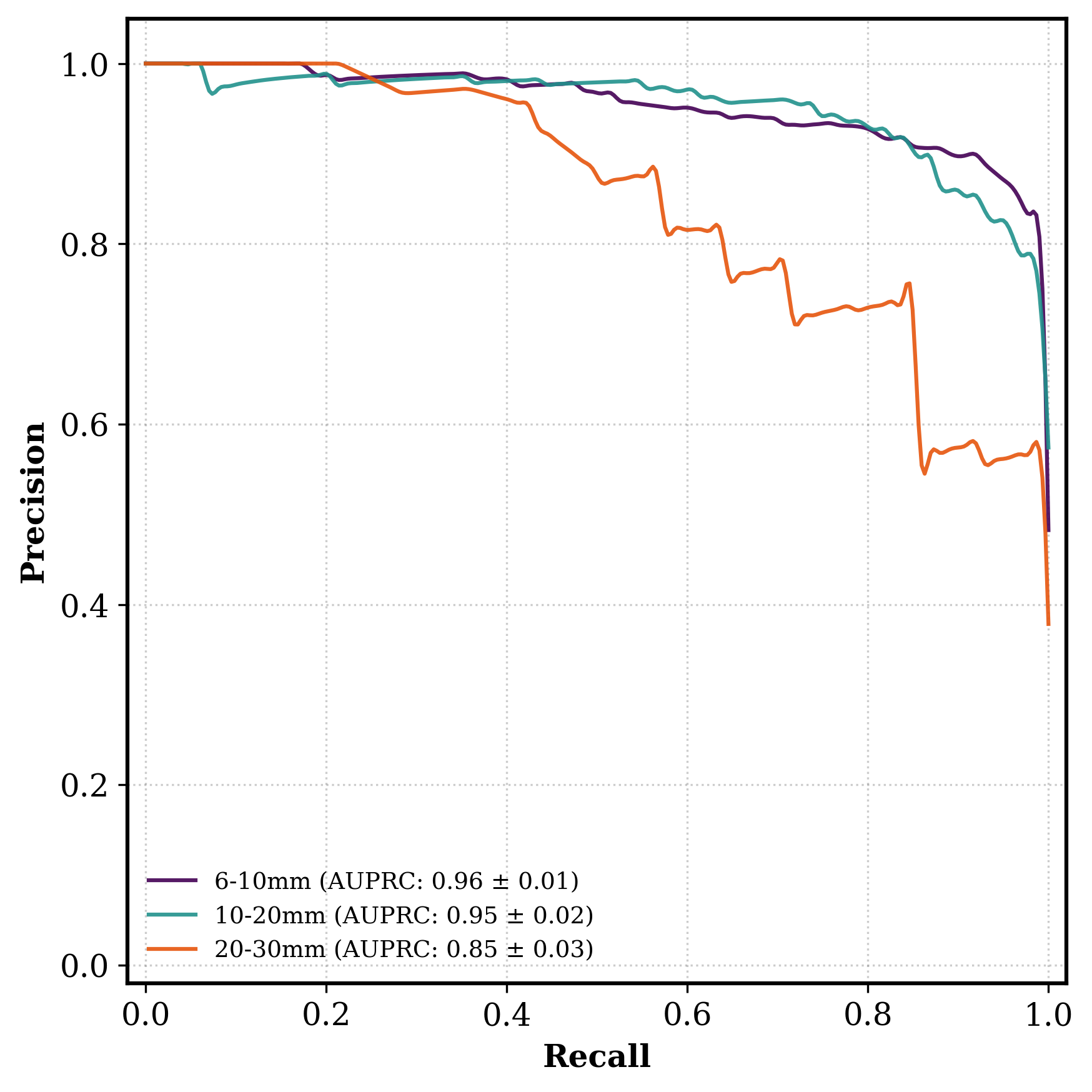} 
        \caption{Ground-glass: \textit{Real} (Baseline)}
        \label{fig:ggo-imbalanced}
    \end{subfigure}
    \hfill
    \begin{subfigure}[b]{0.48\linewidth}
        \centering
        \includegraphics[width=\linewidth]{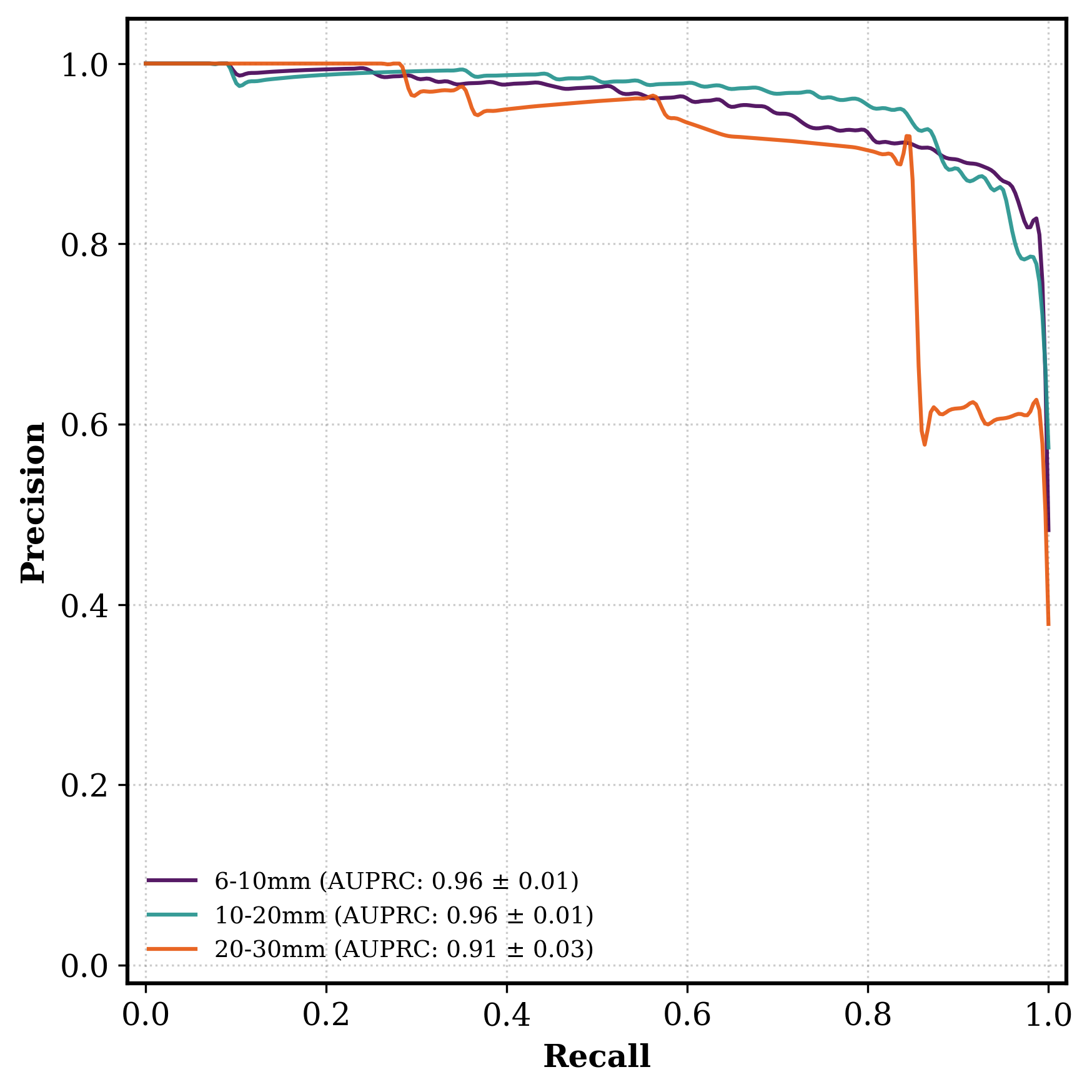} 
        \caption{Ground-glass: \textit{Real + Synth} (Ours)}
        \label{fig:ggo-diffusion}
    \end{subfigure}

    \caption{Precision-Recall curves (AUPRC) comparing baseline against synthetically augmented model. The top row illustrates performance for part-solid nodules, and the bottom row displays ground-glass nodules.}
    \label{fig:auprc_comparison}
\end{figure}

Next, for smaller minority nodules (6-10mm in part-solid and ground-glass), we see that the AUROC differences between the baseline and the proposed method were statistically insignificant across all subtypes. However, we note that the AUPRC for the highly imbalanced part-solid class in the 6-10mm range showed a statistically significant early improvement, increasing from $0.332 \pm 0.06$ to $0.373 \pm 0.05$ ($p = 0.049$, Figures \ref{fig:ps-imbalanced} and \ref{fig:ps-diffusion}). This highlights that metrics such as AUPRC have a benefit in quantifying performance for small, challenging lesions that AUROC alone fails to capture.

More importantly, we observe the most substantial benefits of synthetic augmentation in the 10-20mm and 20-30mm ranges for both part-solid and ground-glass nodules. Specifically, in the 10-20mm category, part-solid AUROC improved significantly from $0.753 \pm 0.04$ to $0.802 \pm 0.03$ ($p = 0.047$), and AUPRC rose from $0.493 \pm 0.07$ to $0.570 \pm 0.03$ ($p = 0.032$). Also, ground-glass nodules in this same size range show significant, consistent gains in both metrics with a $p = 0.046$ (Figures \ref{fig:ggo-imbalanced} and \ref{fig:ggo-diffusion}).

Finally, we see that the performance gap widened most for the largest nodules (20-30mm), where our method yielded highly significant improvements. Part-solid AUROC increased from $0.750 \pm 0.03$ to $0.831 \pm 0.03$ ($p = 0.006$), while AUPRC increased from $0.532 \pm 0.04$ to $0.657 \pm 0.05$ ($p = 0.002$, visually evident in the widened gap between the orange curves in Figures \ref{fig:ps-imbalanced} and \ref{fig:ps-diffusion}). Similarly, ground-glass performance improved, with AUROC reaching $0.933 \pm 0.02$ ($p = 0.004$) and AUPRC reaching $0.910 \pm 0.03$ ($p = 0.013$). 

These results confirm that when we perform a targeted data augmentation using the synthetic samples generated by our proposed approach, it effectively addresses class imbalance, providing statistically significant classification improvements for minority subtypes and larger nodule sizes without compromising majority class accuracy. 

\subsection{Ablation Study: Robustness to Data Scarcity}

\begin{figure*}[h!]
    \centering
    \includegraphics[width=\textwidth]{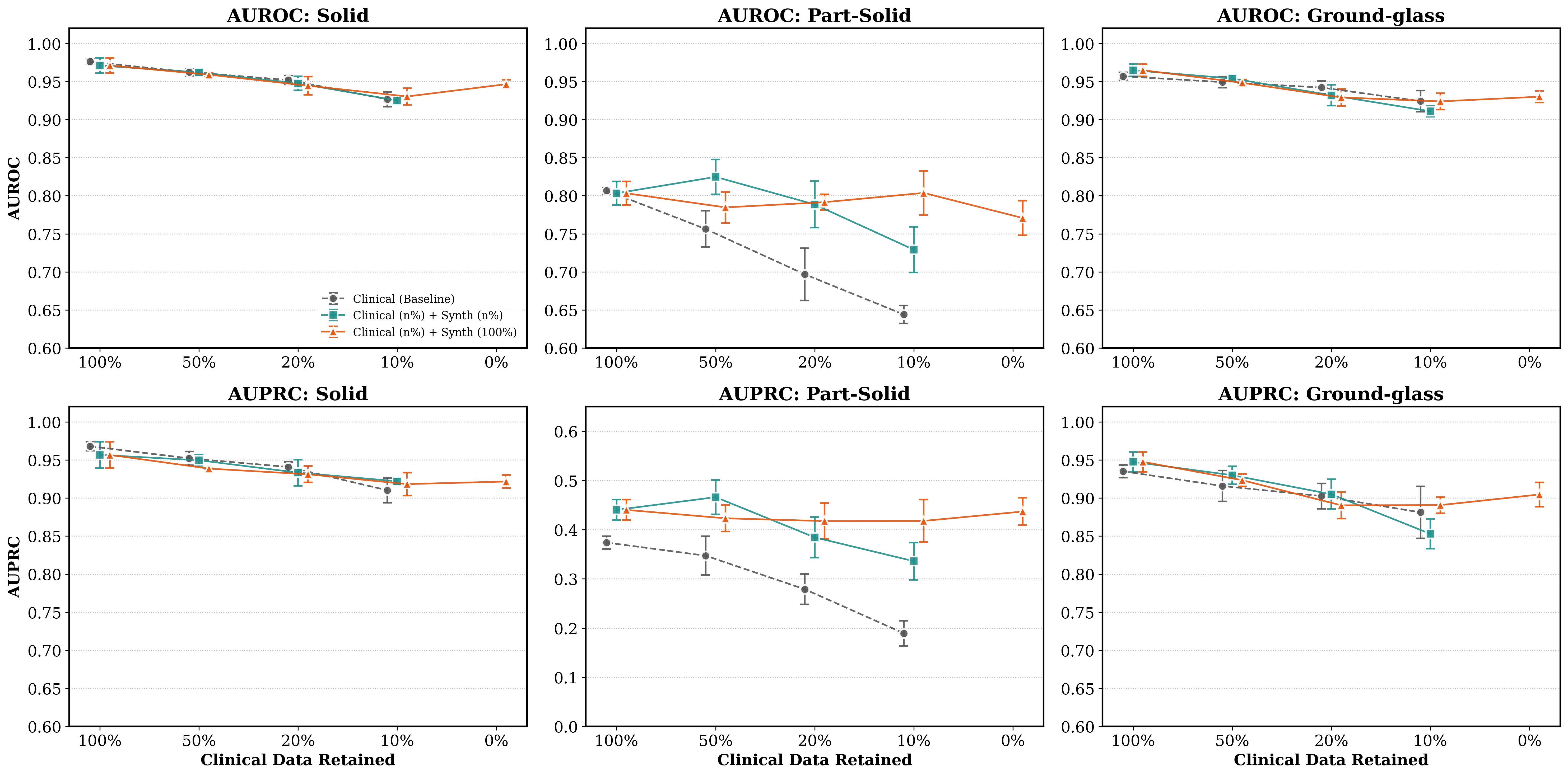}
    \caption{Ablation study evaluating classification robustness under severe data scarcity. Performance metrics (AUROC and AUPRC) are shown across nodule subtypes as the proportion of retained clinical training data is systematically reduced from 100\% to 0\%. The baseline model (\textcolor{Gray}{gray}) is compared against two augmentation strategies: proportional addition of synthetic data (Clinical (n\%) + Synth (n\%), \textcolor{TealBlue}{teal}) and full addition of synthetic data (Clinical (n\%) + Synth (100\%), \textcolor{Orange}{orange}). While well-represented subtypes (solid, ground-glass) are generally resilient to data reduction, the minority part-solid subtype suffers severe baseline degradation. Here, full synthetic augmentation stabilizes part-solid performance even at extreme scarcity levels (10\% and 0\% clinical retention), demonstrating the utility of generated data's features.}
    \label{fig:subtype-ablation}
\end{figure*}

In this section, we investigate the robustness of our framework under conditions of severe data scarcity, where we systematically reduce the amount of available clinical training data and substitute it proportionally with synthetic data for lung nodule subtype classification task. Here, our goal is to assess if the generated data can provide meaningful feature representations to the model in situations of severe real data scarcity. We do so by training 3D DenseNet-121 model across five data retention levels where 100\%, 50\%, 20\%, 10\%, and 0\% of the original clinical dataset is retained as training dataset proportions. Subsequently, we substitute it with generated data in two different augmentation settings: proportional synthetic augmentation (Clinical (n\%) + Synth (n\%)) and full synthetic augmentation (Clinical (n\%) + Synth (100\%)). In proportional synthetic augmentation setting, when we remove a certain proportion of the real clinical data, we add only the same proportion of generated data. For example, in 20\% mark, we use 20\% of clinical data and only use 20\% of generated data; whereas, in full synthetic augmentation setting, when we remove a certain proportion of the real clinical data, we add full synthetic data to the training dataset to make sure that the synthetic samples are disproportionately higher than the real clinical data. The latter setting evaluates whether adding a larger proportion of synthetic samples adversely affects model performance. 

Figure \ref{fig:subtype-ablation} shows the performance of the model across different retention levels quantified using AUROC and AUPRC. For the well-represented solid and the second well represented ground-glass nodule subtypes, we see that the baseline model exhibits high resilience to data reduction. For example, as we reduce the clinical data to 10\%, we see that the baseline AUROC and AUPRC remained relatively stable, only degrading slightly. The addition of synthetic data in these categories maintained consistent results with the baseline, showing that the synthetic samples do not introduce detrimental noise into already easily learnable classes.

Conversely, the minority part-solid subtype demonstrated extreme sensitivity to data reduction. The baseline model's performance deteriorated rapidly as clinical data was removed, with AUROC dropping from approximately 0.80 at 100\% retention to 0.64 at 10\% retention. The AUPRC baseline suffered a similarly steep decline, falling from roughly 0.38 to below 0.20.

The introduction of synthetic data substantially mitigated this degradation. Supplementing the reduced datasets with a proportional amount of synthetic data (Synth (n\%)) delayed the performance drop, maintaining higher accuracy down to the 50\% retention mark before declining. More importantly, supplementing the scarce clinical data with a full complement of synthetic data (Synth (100\%)) completely stabilized the model. For the part-solid class, the Synth (100\%) strategy maintained a consistent AUROC near 0.80 and an AUPRC above 0.40 across all data retention levels, even when utilizing only 10\% or 0\% of the original clinical data. 

These ablation results demonstrate that the proposed method generates highly informative features capable of replacing missing clinical data. The synthetic augmentation acts as a counterbalance for minority classes, preserving diagnostic performance in extreme data-scarce regimes.

\subsection{Malignancy Classification}

\begin{figure}[!t]
    \centering
    \includegraphics[width=\linewidth]{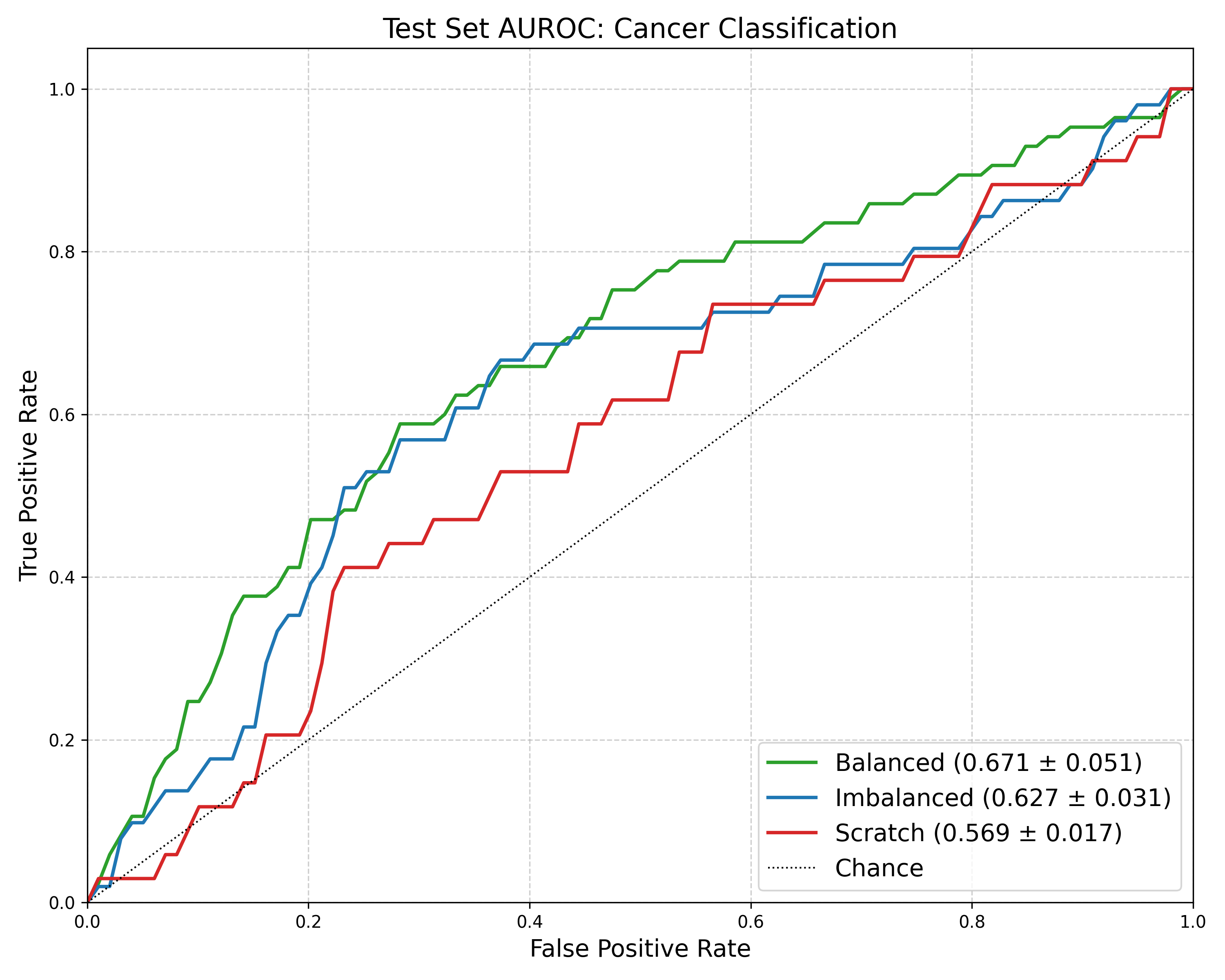}
    \caption{AUROC curves evaluating downstream lung nodule malignancy classification (malignant vs. benign). The plot compares the predictive performance of a 3D DenseNet-121 classifier across three training configurations: trained from scratch on DLCS dataset (\textcolor{Red}{red}), model weights initialized with the subtype classification model trained using imbalanced data and fine-tuned on DLCS (\textcolor{RoyalBlue}{blue}), and model weights initialized with the subtype classification model trained using imbalanced and synthetic data balanced using our proposed generative framework and fine-tuned (\textcolor{Green}{green}). The synthetically balanced model achieves the highest discrimination showing that correcting dataset imbalance with generated data measurably improves diagnostic classification over standard baselines.}
    \label{fig:dlcs_performance}
\end{figure}

In this section, we evaluate the performance of our approach for the lung nodule malignancy classification task. We employ a 3D DenseNet-121 architecture adapted for binary classification (malignant versus benign) and assess performance on a held-out test set using the AUROC metric. We compare three training configurations of this network: (i) a model trained from scratch on the original DLCS dataset, (ii) a model initialized from subtype classification weights trained on the imbalanced dataset and subsequently fine-tuned for malignancy prediction, and (iii) a model initialized from subtype classification weights trained on the synthetically balanced dataset and fine-tuned in the same manner.

The model trained from scratch yielded the lowest discrimination, achieving an AUROC of $0.569 \pm 0.017$, performing only marginally better than random chance (Figure \ref{fig:dlcs_performance}). Initializing the network with subtype classification weights trained on the imbalanced dataset improves predictive performance, resulting in an AUROC of $0.627 \pm 0.031$. 

The highest performance is achieved by the model initialized from subtype classification weights learned on the synthetically balanced dataset. This configuration attains an AUROC of $0.671 \pm 0.051$, demonstrating a clear improvement over both the scratch and imbalanced baselines. 

These results indicate that subtype-informed pretraining provides a strong initialization for malignancy classification, and that incorporating synthetic nodules to balance the training distribution further enhances the quality of learned representations, leading to improved discrimination between benign and malignant lesions.

\section{Discussion}

The development of robust computer-aided diagnosis systems for lung cancer screening is fundamentally constrained by the scarcity and imbalance of annotated data, particularly for clinically important but underrepresented nodule subtypes. In this work, we address this limitation by introducing a controllable generative framework that not only produces realistic pulmonary nodules but also generates samples that are functionally useful for downstream learning tasks.

A key insight of this study is that realism in medical image synthesis is not solely a structural problem, but also a distributional one. While prior generative approaches primarily optimize for visual fidelity through reconstruction or adversarial objectives, they often fail to capture the subtle intensity distributions that distinguish clinically relevant subtypes such as part-solid and ground-glass nodules. The proposed histogram-based regularization explicitly addresses this gap by enforcing alignment in intensity distributions, thereby guiding the model toward subtype-consistent attenuation patterns. This provides a more clinically meaningful notion of realism, beyond visual plausibility alone.

In addition, the use of multi-modal conditioning enables a more disentangled and controllable generative process. By separating semantic subtype, spatial morphology, and intensity characteristics into distinct conditioning signals, the model is able to independently control different attributes of the synthesized nodules. This structured conditioning, combined with lesion-focused latent inpainting, allows the model to generate localized pathological changes while preserving the surrounding anatomical context. Such locality is important in medical imaging, where global image synthesis can easily introduce anatomically inconsistent artifacts.

Another important implication of this work is that synthetic data can serve not only as a means of augmentation, but as a mechanism for representation learning. The observed improvements in downstream classification tasks suggest that the generated samples capture meaningful feature variations that are underrepresented in real datasets. In particular, the combination of subtype-informed pretraining and synthetic balancing indicates that generative models can help reshape the training distribution in a way that improves feature transferability across related clinical tasks, such as malignancy prediction. These findings are especially relevant in the context of data scarcity, where acquiring large, well-balanced annotated datasets is often impractical. The ability of the proposed framework to stabilize model performance under reduced data availability suggests that generative augmentation can act as a substitute for missing clinical diversity, rather than merely a supplement. This has important implications for developing robust learning systems in low-resource settings or rare disease scenarios.

From a clinical perspective, the ability to generate anatomically consistent and subtype-specific nodules offers a pathway toward improving the robustness and fairness of diagnostic models. By augmenting underrepresented classes without degrading performance on majority classes, the proposed approach may help reduce bias in computer-aided diagnosis systems and improve sensitivity to clinically subtle lesion types.

Despite these contributions, several limitations remain. The evaluation of synthetic realism, while supported by quantitative metrics and expert assessment, would benefit from larger multi-reader studies to establish clinical reliability. Furthermore, the dependence on segmentation masks and subtype annotations introduces an additional requirement for labeled data, which may limit scalability in certain settings. In addition, the current framework employs a downsampled segmentation mask as a spatial conditioning signal, which may not fully capture fine-grained lesion boundary information; incorporating higher-resolution spatial conditioning strategies remains an important direction for future work. Finally, the computational complexity of 3D diffusion models remains a practical challenge for widespread deployment.

Future work will focus on addressing these limitations by exploring annotation efficient training strategies, expanding validation across diverse multi-center datasets, and integrating the framework into end-to-end clinical pipelines. 
Extending the approach to other thoracic abnormalities or multi-lesion scenarios may further enhance its clinical applicability.

\section{Conclusion}

In this work, we presented HR-LDM, a histogram-regularized latent diffusion framework for controllable lung nodule synthesis. By integrating multi-modal conditioning with a novel histogram-based regularization strategy and lesion-focused inpainting, the proposed method generates anatomically consistent and visually realistic nodules that closely match the distribution of real clinical data. Extensive experiments demonstrate that the generated samples not only achieve superior fidelity compared to existing methods but also provide tangible benefits for downstream tasks, including improved subtype classification, enhanced robustness under data scarcity, and improved malignancy prediction through subtype-informed pretraining. Overall, this work highlights the potential of diffusion-based generative models as a powerful tool for addressing data limitations in medical imaging, offering a pathway toward more robust and generalizable computer-aided diagnosis systems.

\section{Disclaimer}

The concepts and information presented in this paper are based on research results that are not commercially available. Future commercial availability cannot be guaranteed.

\bibliographystyle{elsarticle-harv} 
\bibliography{references}

@article{bray2024global,
  title={Global cancer statistics 2022: GLOBOCAN estimates of incidence and mortality worldwide for 36 cancers in 185 countries},
  author={Bray, Freddie and Laversanne, Mathieu and Sung, Hyuna and Ferlay, Jacques and Siegel, Rebecca L and Soerjomataram, Isabelle and Jemal, Ahmedin},
  journal={CA: a cancer journal for clinicians},
  volume={74},
  number={3},
  pages={229--263},
  year={2024},
  publisher={Wiley Online Library}
}

@article{national2011reduced,
  title={Reduced lung-cancer mortality with low-dose computed tomographic screening},
  author={National Lung Screening Trial Research Team},
  journal={New England Journal of Medicine},
  volume={365},
  number={5},
  pages={395--409},
  year={2011},
  publisher={Mass Medical Soc}
}

@article{de2020reduced,
  title={Reduced lung-cancer mortality with volume CT screening in a randomized trial},
  author={de Koning, Harry J and van Der Aalst, Carlijn M and de Jong, Pim A and Scholten, Ernst T and Nackaerts, Kristiaan and Heuvelmans, Marjolein A and Lammers, Jan-Willem J and Weenink, Carla and Yousaf-Khan, Uraujh and Horeweg, Nanda and others},
  journal={New England journal of medicine},
  volume={382},
  number={6},
  pages={503--513},
  year={2020},
  publisher={Mass Medical Soc}
}

@article{landy2021using,
  title={Using prediction models to reduce persistent racial and ethnic disparities in the draft 2020 USPSTF lung cancer screening guidelines},
  author={Landy, Rebecca and Young, Corey D and Skarzynski, Martin and Cheung, Li C and Berg, Christine D and Rivera, M Patricia and Robbins, Hilary A and Chaturvedi, Anil K and Katki, Hormuzd A},
  journal={JNCI: Journal of the National Cancer Institute},
  volume={113},
  number={11},
  pages={1590--1594},
  year={2021},
  publisher={Oxford University Press}
}

@article{henschke2002ct,
  title={CT screening for lung cancer: frequency and significance of part-solid and nonsolid nodules},
  author={Henschke, Claudia I and Yankelevitz, David F and Mirtcheva, Rosna and McGuinness, Georgeann and McCauley, Dorothy and Miettinen, Olli S},
  journal={American Journal of Roentgenology},
  volume={178},
  number={5},
  pages={1053--1057},
  year={2002},
  publisher={American Roentgen Ray Society}
}

@article{hammer2019cancer,
  title={Cancer risk in subsolid nodules in the National Lung Screening Trial},
  author={Hammer, Mark M and Palazzo, Lauren L and Kong, Chung Yin and Hunsaker, Andetta R},
  journal={Radiology},
  volume={293},
  number={2},
  pages={441--448},
  year={2019},
  publisher={Radiological Society of North America}
}

@article{national2011national,
  title={The national lung screening trial: overview and study design},
  author={National Lung Screening Trial Research Team},
  journal={Radiology},
  volume={258},
  number={1},
  pages={243--253},
  year={2011},
  publisher={Radiological Society of North America, Inc.}
}

@article{armato2011lung,
  title={The lung image database consortium (LIDC) and image database resource initiative (IDRI): a completed reference database of lung nodules on CT scans},
  author={Armato III, Samuel G and McLennan, Geoffrey and Bidaut, Luc and McNitt-Gray, Michael F and Meyer, Charles R and Reeves, Anthony P and Zhao, Binsheng and Aberle, Denise R and Henschke, Claudia I and Hoffman, Eric A and others},
  journal={Medical physics},
  volume={38},
  number={2},
  pages={915--931},
  year={2011},
  publisher={Wiley Online Library}
}

@inproceedings{yang2019class,
  title={Class-aware adversarial lung nodule synthesis in CT images},
  author={Yang, Jie and Liu, Siqi and Grbic, Sasa and Setio, Arnaud Arindra Adiyoso and Xu, Zhoubing and Gibson, Eli and Chabin, Guillaume and Georgescu, Bogdan and Laine, Andrew F and Comaniciu, Dorin},
  booktitle={2019 IEEE 16th International Symposium on Biomedical Imaging (ISBI 2019)},
  pages={1348--1352},
  year={2019},
  organization={IEEE}
}

@article{wang2021realistic,
  title={Realistic lung nodule synthesis with multi-target co-guided adversarial mechanism},
  author={Wang, Qiuli and Zhang, Xiaohong and Zhang, Wei and Gao, Mingchen and Huang, Sheng and Wang, Jian and Zhang, Jiuquan and Yang, Dan and Liu, Chen},
  journal={IEEE Transactions on Medical Imaging},
  volume={40},
  number={9},
  pages={2343--2353},
  year={2021},
  publisher={IEEE}
}

@inproceedings{guo2025maisi,
  title={Maisi: Medical ai for synthetic imaging},
  author={Guo, Pengfei and Zhao, Can and Yang, Dong and Xu, Ziyue and Nath, Vishwesh and Tang, Yucheng and Simon, Benjamin and Belue, Mason and Harmon, Stephanie and Turkbey, Baris and others},
  booktitle={2025 IEEE/CVF Winter Conference on Applications of Computer Vision (WACV)},
  pages={4430--4441},
  year={2025},
  organization={IEEE}
}

@inproceedings{chen2024towards,
  title={Towards generalizable tumor synthesis},
  author={Chen, Qi and Chen, Xiaoxi and Song, Haorui and Xiong, Zhiwei and Yuille, Alan and Wei, Chen and Zhou, Zongwei},
  booktitle={Proceedings of the IEEE/CVF conference on computer vision and pattern recognition},
  pages={11147--11158},
  year={2024}
}

@inproceedings{lei2025lesiondiffusion,
  title={LesionDiffusion: Towards Text-Controlled General Lesion Synthesis},
  author={Lei, Wenhui and Tian, Hengrui and Dai, Linrui and Chen, Hanyu and Zhang, Xiaofan},
  booktitle={International Conference on Medical Image Computing and Computer-Assisted Intervention},
  pages={327--336},
  year={2025},
  organization={Springer}
}

@inproceedings{lefusion,
  author       = {Hantao Zhang and
                  Yuhe Liu and
                  Jiancheng Yang and
                  Shouhong Wan and
                  Xinyuan Wang and
                  Wei Peng and
                  Pascal Fua},
  title        = {LeFusion: Controllable Pathology Synthesis via Lesion-Focused Diffusion
                  Models},
  booktitle    = {The Thirteenth International Conference on Learning Representations,
                  {ICLR} 2025, Singapore, April 24-28, 2025},
  year         = {2025},
  bibsource    = {dblp computer science bibliography, https://dblp.org}
}

@article{goodfellow2020generative,
  title={Generative adversarial networks},
  author={Goodfellow, Ian and Pouget-Abadie, Jean and Mirza, Mehdi and Xu, Bing and Warde-Farley, David and Ozair, Sherjil and Courville, Aaron and Bengio, Yoshua},
  journal={Communications of the ACM},
  volume={63},
  number={11},
  pages={139--144},
  year={2020},
  publisher={ACM New York, NY, USA}
}

@article{shin2018abnormal,
  title={Abnormal colon polyp image synthesis using conditional adversarial networks for improved detection performance},
  author={Shin, Younghak and Qadir, Hemin Ali and Balasingham, Ilangko},
  journal={IEEE Access},
  volume={6},
  pages={56007--56017},
  year={2018},
  publisher={IEEE}
}

@inproceedings{frid2018synthetic,
  title={Synthetic data augmentation using GAN for improved liver lesion classification},
  author={Frid-Adar, Maayan and Klang, Eyal and Amitai, Michal and Goldberger, Jacob and Greenspan, Hayit},
  booktitle={2018 IEEE 15th international symposium on biomedical imaging (ISBI 2018)},
  pages={289--293},
  year={2018},
  organization={IEEE}
}

@inproceedings{han2018gan,
  title={GAN-based synthetic brain MR image generation},
  author={Han, Changhee and Hayashi, Hideaki and Rundo, Leonardo and Araki, Ryosuke and Shimoda, Wataru and Muramatsu, Shinichi and Furukawa, Yujiro and Mauri, Giancarlo and Nakayama, Hideki},
  booktitle={2018 IEEE 15th international symposium on biomedical imaging (ISBI 2018)},
  pages={734--738},
  year={2018},
  organization={IEEE}
}

@inproceedings{han2019synthesizing,
  title={Synthesizing diverse lung nodules wherever massively: 3D multi-conditional GAN-based CT image augmentation for object detection},
  author={Han, Changhee and Kitamura, Yoshiro and Kudo, Akira and Ichinose, Akimichi and Rundo, Leonardo and Furukawa, Yujiro and Umemoto, Kazuki and Li, Yuanzhong and Nakayama, Hideki},
  booktitle={2019 International Conference on 3D Vision (3DV)},
  pages={729--737},
  year={2019},
  organization={IEEE}
}

@article{jin2021free,
  title={Free-form tumor synthesis in computed tomography images via richer generative adversarial network},
  author={Jin, Qiangguo and Cui, Hui and Sun, Changming and Meng, Zhaopeng and Su, Ran},
  journal={Knowledge-Based Systems},
  volume={218},
  pages={106753},
  year={2021},
  publisher={Elsevier}
}

@article{ho2020denoising,
  title={Denoising diffusion probabilistic models},
  author={Ho, Jonathan and Jain, Ajay and Abbeel, Pieter},
  journal={Advances in neural information processing systems},
  volume={33},
  pages={6840--6851},
  year={2020}
}

@inproceedings{rombach2022high,
  title={High-resolution image synthesis with latent diffusion models},
  author={Rombach, Robin and Blattmann, Andreas and Lorenz, Dominik and Esser, Patrick and Ommer, Bj{\"o}rn},
  booktitle={Proceedings of the IEEE/CVF conference on computer vision and pattern recognition},
  pages={10684--10695},
  year={2022}
}

@inproceedings{zhang2023adding,
  title={Adding conditional control to text-to-image diffusion models},
  author={Zhang, Lvmin and Rao, Anyi and Agrawala, Maneesh},
  booktitle={Proceedings of the IEEE/CVF international conference on computer vision},
  pages={3836--3847},
  year={2023}
}

@article{liu2020no,
  title={No surprises: Training robust lung nodule detection for low-dose CT scans by augmenting with adversarial attacks},
  author={Liu, Siqi and Setio, Arnaud Arindra Adiyoso and Ghesu, Florin C and Gibson, Eli and Grbic, Sasa and Georgescu, Bogdan and Comaniciu, Dorin},
  journal={IEEE Transactions on Medical Imaging},
  volume={40},
  number={1},
  pages={335--345},
  year={2020},
  publisher={IEEE}
}

@article{hendrix2023deep,
  title={Deep learning for the detection of benign and malignant pulmonary nodules in non-screening chest CT scans},
  author={Hendrix, Ward and Hendrix, Nils and Scholten, Ernst T and Mourits, Mari{\"e}lle and Trap-de Jong, Joline and Schalekamp, Steven and Korst, Mike and Van Leuken, Maarten and Van Ginneken, Bram and Prokop, Mathias and others},
  journal={Communications medicine},
  volume={3},
  number={1},
  pages={156},
  year={2023},
  publisher={Nature Publishing Group UK London}
}

@article{wang2025duke,
  title={The Duke Lung Cancer Screening (DLCS) dataset: a reference dataset of annotated low-dose screening thoracic CT},
  author={Wang, Avivah J and Tushar, Fakrul Islam and Harowicz, Michael R and Tong, Betty C and Lafata, Kyle J and Tailor, Tina D and Lo, Joseph Y},
  journal={Radiology: Artificial Intelligence},
  volume={7},
  number={4},
  pages={e240248},
  year={2025},
  publisher={Radiological Society of North America}
}

@inproceedings{esser2021taming,
  title={Taming transformers for high-resolution image synthesis},
  author={Esser, Patrick and Rombach, Robin and Ommer, Bjorn},
  booktitle={Proceedings of the IEEE/CVF conference on computer vision and pattern recognition},
  pages={12873--12883},
  year={2021}
}

@inproceedings{zhang2018unreasonable,
  title={The unreasonable effectiveness of deep features as a perceptual metric},
  author={Zhang, Richard and Isola, Phillip and Efros, Alexei A and Shechtman, Eli and Wang, Oliver},
  booktitle={Proceedings of the IEEE conference on computer vision and pattern recognition},
  pages={586--595},
  year={2018}
}

@inproceedings{
yu2022vectorquantized,
title={Vector-quantized Image Modeling with Improved {VQGAN}},
author={Jiahui Yu and Xin Li and Jing Yu Koh and Han Zhang and Ruoming Pang and James Qin and Alexander Ku and Yuanzhong Xu and Jason Baldridge and Yonghui Wu},
booktitle={International Conference on Learning Representations},
year={2022},
url={https://openreview.net/forum?id=pfNyExj7z2}
}

@inproceedings{
berrada2025boosting,
title={Boosting Latent Diffusion with Perceptual Objectives},
author={Tariq Berrada and Pietro Astolfi and Melissa Hall and Marton Havasi and Yohann Benchetrit and Adriana Romero-Soriano and Karteek Alahari and Michal Drozdzal and Jakob Verbeek},
booktitle={The Thirteenth International Conference on Learning Representations},
year={2025},
url={https://openreview.net/forum?id=y4DtzADzd1}
}

@article{ustinova2016learning,
  title={Learning deep embeddings with histogram loss},
  author={Ustinova, Evgeniya and Lempitsky, Victor},
  journal={Advances in neural information processing systems},
  volume={29},
  year={2016}
}

@article{lin2002divergence,
  title={Divergence measures based on the Shannon entropy},
  author={Lin, Jianhua},
  journal={IEEE Transactions on Information theory},
  volume={37},
  number={1},
  pages={145--151},
  year={2002},
  publisher={IEEE}
}

@inproceedings{nichol2021improved,
  title={Improved denoising diffusion probabilistic models},
  author={Nichol, Alexander Quinn and Dhariwal, Prafulla},
  booktitle={International conference on machine learning},
  pages={8162--8171},
  year={2021},
  organization={PMLR}
}

@inproceedings{saharia2022palette,
  title={Palette: Image-to-image diffusion models},
  author={Saharia, Chitwan and Chan, William and Chang, Huiwen and Lee, Chris and Ho, Jonathan and Salimans, Tim and Fleet, David and Norouzi, Mohammad},
  booktitle={ACM SIGGRAPH 2022 conference proceedings},
  pages={1--10},
  year={2022}
}

@inproceedings{ronneberger2015u,
  title={U-net: Convolutional networks for biomedical image segmentation},
  author={Ronneberger, Olaf and Fischer, Philipp and Brox, Thomas},
  booktitle={International Conference on Medical image computing and computer-assisted intervention},
  pages={234--241},
  year={2015},
  organization={Springer}
}

@article{kingma2013auto,
  title={Auto-encoding variational bayes},
  author={Kingma, Diederik P and Welling, Max},
  journal={arXiv preprint arXiv:1312.6114},
  year={2013}
}

@article{wasserthal2023totalsegmentator,
  title={TotalSegmentator: robust segmentation of 104 anatomic structures in CT images},
  author={Wasserthal, Jakob and Breit, Hanns-Christian and Meyer, Manfred T and Pradella, Maurice and Hinck, Daniel and Sauter, Alexander W and Heye, Tobias and Boll, Daniel T and Cyriac, Joshy and Yang, Shan and others},
  journal={Radiology: Artificial Intelligence},
  volume={5},
  number={5},
  pages={e230024},
  year={2023},
  publisher={Radiological Society of North America}
}

@article{isensee2021nnu,
  title={nnU-Net: a self-configuring method for deep learning-based biomedical image segmentation},
  author={Isensee, Fabian and Jaeger, Paul F and Kohl, Simon AA and Petersen, Jens and Maier-Hein, Klaus H},
  journal={Nature methods},
  volume={18},
  number={2},
  pages={203--211},
  year={2021},
  publisher={Nature Publishing Group US New York}
}

@article{cardoso2022monai,
  title={Monai: An open-source framework for deep learning in healthcare},
  author={Cardoso, M Jorge and Li, Wenqi and Brown, Richard and Ma, Nic and Kerfoot, Eric and Wang, Yiheng and Murrey, Benjamin and Myronenko, Andriy and Zhao, Can and Yang, Dong and others},
  journal={arXiv preprint arXiv:2211.02701},
  year={2022}
}

@article{mei2022radimagenet,
  title={RadImageNet: an open radiologic deep learning research dataset for effective transfer learning},
  author={Mei, Xueyan and Liu, Zelong and Robson, Philip M and Marinelli, Brett and Huang, Mingqian and Doshi, Amish and Jacobi, Adam and Cao, Chendi and Link, Katherine E and Yang, Thomas and others},
  journal={Radiology: Artificial Intelligence},
  volume={4},
  number={5},
  pages={e210315},
  year={2022},
  publisher={Radiological Society of North America}
}

@inproceedings{huang2017densely,
  title={Densely connected convolutional networks},
  author={Huang, Gao and Liu, Zhuang and Van Der Maaten, Laurens and Weinberger, Kilian Q},
  booktitle={Proceedings of the IEEE conference on computer vision and pattern recognition},
  pages={4700--4708},
  year={2017}
}

@inproceedings{cciccek20163d,
  title={3D U-Net: learning dense volumetric segmentation from sparse annotation},
  author={{\c{C}}i{\c{c}}ek, {\"O}zg{\"u}n and Abdulkadir, Ahmed and Lienkamp, Soeren S and Brox, Thomas and Ronneberger, Olaf},
  booktitle={International conference on medical image computing and computer-assisted intervention},
  pages={424--432},
  year={2016},
  organization={Springer}
}

@article{mortani2026deep,
  title={Deep learning-based pulmonary nodule risk assessment outperforms established malignancy risk scores in lung cancer screening},
  author={Mortani Barbosa Jr, Eduardo J and Kim, Yohan and Zhang, Yanbo and Setio, Arnaud AA and Mellot, Francois and Grenier, Philippe A and Zimmermann, Mathis and Georgescu, Bogdan and Grbic, Sasa and Gefter, Warren B},
  journal={Radiology Advances},
  volume={3},
  number={1},
  pages={umag003},
  year={2026},
  publisher={Oxford University Press}
}

@misc{nlst2013,
  author       = {{National Lung Screening Trial Research Team}},
  title        = {Data from the National Lung Screening Trial (NLST) [Data set]},
  year         = {2013},
  howpublished = {The Cancer Imaging Archive},  url          = {https://doi.org/10.7937/TCIA.HMQ8-J677}
}

@inproceedings{zhao2026maisi,
  title={Maisi-v2: Accelerated 3d high-resolution medical image synthesis with rectified flow and region-specific contrastive loss},
  author={Zhao, Can and Guo, Pengfei and Yang, Dong and He, Yufan and Tang, Yucheng and Simon, Benjamin and Belue, Mason and Harmon, Stephanie and Turkbey, Baris and Xu, Daguang},
  booktitle={Proceedings of the AAAI Conference on Artificial Intelligence},
  volume={40},
  number={15},
  pages={13088--13098},
  year={2026}
}

\end{document}